\DocumentMetadata{}
\documentclass[sigconf, nonacm]{acmart}

\makeatletter
\def\@ACM@checkaffil{
    \if@ACM@instpresent\else
    \ClassWarningNoLine{\@classname}{No institution present for an affiliation}%
    \fi
    \if@ACM@citypresent\else
    \ClassWarningNoLine{\@classname}{No city present for an affiliation}%
    \fi
    \if@ACM@countrypresent\else
        \ClassWarningNoLine{\@classname}{No country present for an affiliation}%
    \fi
}
\makeatother

\newcommand{\one}{\textbf{({\em i}\/)}\xspace}
\newcommand{\two}{\textbf{({\em ii}\/)}\xspace}
\newcommand{\three}{\textbf{({\em iii}\/)}\xspace}
\newcommand{\four}{\textbf{({\em iv}\/)}\xspace}
\newcommand{\five}{\textbf{({\em v}\/)}\xspace}

\def\eg{\emph{e.g.,}\xspace}

\def\etal{\emph{et al.}\xspace}
\def\vs{\emph{vs.}\xspace}

\newcommand{\pb}[1]{\vspace{0.75ex}\noindent{\bf \em #1}}
\usepackage{xspace}
\usepackage{csquotes}
\usepackage{url}
\usepackage{fontawesome}
\usepackage{tcolorbox}
\usepackage{subcaption}
\usepackage{makecell}
\usepackage{mwe}
\usepackage{multirow}
\usepackage{graphicx}
\usepackage{nicematrix}
\usepackage[normalem]{ulem}
\useunder{\uline}{\ul}{}
\usepackage{fontawesome}
\usepackage{tabularx}
\usepackage{booktabs}
\usepackage{pifont}
\usepackage{multirow}
\usepackage{amsmath}
\usepackage{svg}
\usepackage{xcolor}
\usepackage{enumitem}
\usepackage{array}
\usepackage{diagbox}
\usepackage[ruled,linesnumbered]{algorithm2e}
\newcolumntype{L}[1]{>{\raggedright\let\newline\\\arraybackslash\hspace{0pt}}m{#1}}
\newcolumntype{C}[1]{>{\centering\let\newline\\\arraybackslash\hspace{0pt}}m{#1}}
\newcolumntype{R}[1]{>{\raggedleft\let\newline\\\arraybackslash\hspace{0pt}}m{#1}}

\AtBeginDocument{%
  \providecommand\BibTeX{{%
    \normalfont B\kern-0.5em{\scshape i\kern-0.25em b}\kern-0.8em\TeX}}}

\setcopyright{none}
\copyrightyear{2018}
\acmYear{2018}
\acmDOI{XXXXXXX.XXXXXXX}

\acmConference[Conference acronym 'XX]{Make sure to enter the correct
  conference title from your rights confirmation emai}{June 03--05,
  2018}{Woodstock, NY}
\begin{document}

\title{Exploring the Capability of ChatGPT to Reproduce Human Labels for Social Computing Tasks (Extended Version)}

\author{
    Yiming Zhu,\textsuperscript{\rm *}\quad
    Peixian Zhang,\quad
    Ehsan-Ul Haq,\quad
    Pan Hui,\textsuperscript{\rm *}\\
    Gareth Tyson\textsuperscript{\rm *}\quad
}
\thanks{*~Yiming Zhu is also with The Hong Kong University of Science and Technology, Gareth Tyson is also with Queen Mary University of London, and Pan Hui is also with the University of Helsinki.}
\affiliation{
    \institution{The Hong Kong University of Science and Technology (Guangzhou)\\}
    \{yzhucd, euhaq\}@connect.ust.hk \quad 
    \{panhui, gtyson\}@ust.hk
}

\renewcommand{\shortauthors}{First author and Second author, et al.}

\begin{abstract}
Harnessing the potential of large language models (LLMs) like ChatGPT can help address social challenges through inclusive, ethical, and sustainable means.
In this paper, we investigate the extent to which ChatGPT can annotate data for social computing tasks, aiming to reduce the complexity and cost of undertaking web research. 
To evaluate ChatGPT's potential, we re-annotate seven datasets using ChatGPT, covering topics related to pressing social issues like COVID-19 misinformation, social bot deception, cyberbully, clickbait news, and the Russo-Ukrainian War. Our findings demonstrate that ChatGPT exhibits promise in handling these data annotation tasks, albeit with some challenges. Across the seven datasets, ChatGPT achieves an average annotation F1-score of 72.00\%. Its performance excels in clickbait news annotation, correctly labeling 89.66\% of the data. However, we also observe significant variations in performance across individual labels.
Our study reveals predictable patterns in ChatGPT's annotation performance. 
Thus, we propose GPT-Rater, a tool to predict if ChatGPT can correctly label data for a given annotation task. Researchers can use this to identify where ChatGPT might be suitable for their annotation requirements. We show that GPT-Rater effectively predicts ChatGPT's performance. It performs best on a clickbait headlines dataset by achieving an average F1-score of 95.00\%.
We believe that this research opens new avenues for analysis and can reduce barriers to engaging in social computing research.
\end{abstract}





\maketitle

\section{Introduction}
Crowd-sourced human intelligence is commonly used for text data annotation~\cite{haq2022its,sorokin2009utility,akkaya2010amazon}. 
Such annotations are then used for training various models, including stance detection~\cite{glandt2021stance}, hate speech detection~\cite{he2021racism}, sentiment analysis~\cite{rosenthal2019semeval}, and bot detection~\cite{fagni2021tweepfake}. 
While unsupervised methods are being introduced for classification tasks, such methods usually require large data samples~\cite{usama2019unsupervised, wang2020survey}. Thus, social computing research still relies heavily on human annotations. This, however, creates significant barriers for less well-funded research labs, resulting in a lock-out effect. For example, the annotation of a 10,000-post social media dataset by three human annotators would take approximately five hours. With a rate of \$25 per individual worker, this would cost hundreds of dollars~\cite{snow2008cheap, diaz2022crowdworksheets}. When performing comparative analyses across multiple datasets, these costs can easily escalate to thousands of dollars.



Recently, the release of ChatGPT has uncovered a range of cases where large language models (LLMs) can help substitute human intelligence~\cite{guo2023close, bang2023multitask, zhang2023complete}. Several works compare the use of ChatGPT to human methods~\cite{zhang2022would, sobania2023analysis, wang2023chatgpt}. For instance, researchers have investigated the use of ChatGPT for automatic bug fixing~\cite{sobania2023analysis}, misinformation detection~\cite{sallam2023chatgpt}, and even generating academic writing~\cite{aydin2022openai}.
In this paper, we explore the potential of using ChatGPT for five text-based social computing annotation tasks. We primarily seek to understand whether ChatGPT has the potential to reproduce human-generated annotations. ChatGPT's annotations can highlight its usefulness against crowd-sourced annotations. 
To achieve this, we first use ChatGPT to label seven annotation datasets on five distinct social problems -- COVID-19 controversies (3x), social bot deception, cyberbully, clickbait news, and Russo-Ukrainian War stance detection. 
We compare ChatGPT's labels with the human assigned labels on those datasets. 
Our results show that ChatGPT \emph{does} have the potential to perform data annotation tasks.
Performance is highest for the clickbait headlines dataset, with ChatGPT correctly annotating 89.66\% of headlines. In contrast, performance is worst for the COVID-19 hate speech task, which only correctly annotates 52.24\% of posts. Closer inspection reveals that performance varies substantially across individual labels. We observe significant gaps (over 25\%) exist between labels' accuracy on five out of seven datasets.
For instance, in social bot detection, while ChatGPT manages to identify 81.1\% of human tweets, it can only identify 45.5\% of bot tweets.

Thus, we surmise that researchers must be careful in selecting which tasks they use ChatGPT annotation for.
This raises the question of how researchers can identify which tasks ChatGPT annotations would be suitable for. To address this, we build a tool, GPT-Rater, to predict whether ChatGPT will be able to reproduce the correct label for given text. This allows researchers to estimate how suitable ChatGPT is for their annotation requirements.
Results shows that GPT-Rater effectively predicts the correctness of ChatGPT's annotations. On the clickbait headlines dataset, GPT-Rater manages to predict ChatGPT's annotation correctness with high accuracy ($\mu=90.59\%, \sigma=0.13\%$) and F1-scores ($\mu=95.00\%, \sigma=0.05\%$).
Moreover, for five out of seven datasets, GPT-Rater attains average F1-scores above 75\%, with standard deviations lower than 3\%.
This implies that our tool can effectively give researchers a preview into how suitable ChatGPT automation will be for their annotation tasks. 


We believe this work can open up new lines of analysis and can act as a basis for future research into the use of ChatGPT for human annotation tasks. Our contributions are as follows:

\begin{enumerate}[leftmargin=*]
    \item We evaluate the efficacy of ChatGPT at replacing human annotators for five important social computing data tasks, spanning seven datasets.

    \item We show that ChatGPT \emph{can} replace human annotators, yet this varies across tasks and datasets. Performance is highest for a clickbait headlines detection dataset (89.66\% accuracy), and lowest for hate speech detection (52.24\% accuracy).

    \item We propose GPT-Rater, a tool that estimates how suitable ChatGPT is as an annotator for a given task. GPT-Rater demonstrates high effectiveness, with its best performance observed for the clickbait detection task, achieving 90.59\% accuracy and 95.00\% F1-score. GPT-Rater achieves average F1-scores exceeding 75\% for five out of seven datasets. GPT-Rater exhibits robust performance even on a subset of annotated data, showing its potential to assist researchers in estimating ChatGPT's annotation capability with just a small number of human labels.
    
\end{enumerate}

\section{Related Work} \label{sec:related_work}
Recent research has looked at using ChatGPT for annotating misinformation~\cite{bang2023multitask, sallam2023chatgpt} and hate speech~\cite{huang2023chatgpt}.
Huang \etal~\cite{huang2023chatgpt} report that ChatGPT is able to correctly annotate 80\% of the implicit hateful tweets from the \texttt{LatentHatred} dataset~\cite{elsherief2021latent}.
In addition, the authors show that ChatGPT's explanations can reinforce human annotators' perception of the target text in explaining why tweets would be annotated as hateful or not. 
Our study also examines how well ChatGPT performs in annotating hate speech. However, we do not limit the annotation to a binary decision for whether tweets are hateful or not, and further include a neutral label.
Note, existing literature has highlighted the importance of this, stating that tweets with neutral expressions can act as a defense against the spread of hateful content~\cite{mathew2019thou, schieb2016governing}.

Similar to us, others have used ChatGPT for performing stance detection~\cite{zhang2022would, aiyappa2023can}. 
Zhang \etal~\cite{zhang2022would} evaluate ChatGPT's performance on detecting political stance on two prevalent datasets, \texttt{SemEval-2016}~\cite{mohammad2016semeval} and \texttt{P-Stance}~\cite{li2021p}.
The authors report that ChatGPT can outperform most state-of-art stance detection models in zero-shot settings, suggesting ChatGPT's potential to handle stance annotation tests.
Major challenges still remain though.
Aiyappa \etal~\cite{aiyappa2023can} show that ChatGPT's performance varies across different model versions. The authors also point out that such variance is due to the possibility of data leakage, where past prompts are used for training the next ChatGPT generation.
Nonetheless, it remains unclear how well ChatGPT performs in annotating individuals' stances in a broader context, such as those who are in favor or against a particular issue.
Our analysis also finds further challenges, \eg that ChatGPT has a tendency to overestimate neutral stances.

In addition to the aforementioned themes, others have experimented with ChatGPT to perform tasks such as genre identification~\cite{kuzman2023chatgpt}, topic identification~\cite{gilardi2023chatgpt}, sentiment analysis~\cite{bang2023multitask}, and fake news detection~\cite{bang2023multitask, hoes2023using}. 
However, most of this literature only focuses on a single annotation task. 
In contrast, we seek to perform a comparative analysis across different data annotation tasks aiming to solve social problems. We further propose a tool, GPT-Rater, that can predict if ChatGPT will correctly annotate a given post.





\section{Methodology} \label{sec:method}

\subsection{Overview}

We follow a comparative approach to analyze the differences in human annotations and ChatGPT annotations by utilizing seven different datasets. 
We select seven annotation datasets covering five social problems that are commonly used in academic research: \one COVID-19 controversies (vaccine arguments, anti-Asian hate speech, and COVID-19 fake news)~\cite{poddar2022winds, he2021racism, patwa2020fighting}, \two social bot deception~\cite{alharbi2021social}, \three cyberbully~\cite{kennedy2020constructing}, \four clickbait news~\cite{chakraborty2016stop}, and \five Russo-Ukrainian War~\cite{zhu2022reddit, haq2022twitter, RU-echo}. 
For these seven datasets, we then attempt to recreate the human annotations using ChatGPT. In the following subsections, we describe our datasets (\S\ref{sec:dataset}), approaches for performing ChatGPT annotation (\S\ref{sec:chatgpt_annotation}). 

\subsection{Datasets}\label{sec:dataset}

\begin{table}[t]
\centering
\resizebox{\columnwidth}{!}{%
\begin{tabular}{|L{8em}R{2.6em}|L{5em}R{15em}|}
\toprule
\textbf{Dataset} &
  \textbf{Size} &
  \textbf{ChatGPT annotation} &
  \textbf{Labels (Human/ChatGPT)} \\ \midrule
\textbf{Vaccine Stance} &
  5,926 &
  5,832 \newline (98.41\%) &
  Anti-vaccine (21.56\%/10.62\%)\newline Pro-vaccine (39.72\%/26.86\%)\newline Neutral (38.72\%/62.52\%) \\ \midrule
\textbf{COVID-19 Hate Speech} &
  2,290 &
  2,280 \newline (99.56\%) &
  Hate (18.73\%/34.30\%) \newline Counterhate (22.58\%/23.73\%) \newline Neutral (58.69\%/41.97\%) \\ \midrule
\textbf{COVID-19 Fake News} &
  10,700 & 10,540 \newline (98.50\%)
  &
  Real news (52.34\%/64.00\%) \newline Fake news (47.66\%/36.00\%) \\ \midrule
\textbf{Social Bot} &
  16,824 & 16,235 \newline (96.50\%)
  &
  Human (54.08\%/69.01\%) \newline Bot (45.92\%/30.99\%) \\ \midrule
\textbf{Anti-LGBT Cyberbullying} &
  4,299 & 4,193 \newline (97.53\%)
  &
  Cyberbully (29.22\%/49.10\%) \newline Not cyberbully (70.78\%/50.90\%) \\ \midrule
\textbf{Clickbait Headlines} &
  32,000 & 31,084 \newline (97.14\%)
  &
  Clickbait (50.00\%/40.54\%) \newline Not clickbait (50.00\%/59.46\%) \\ \midrule
\textbf{Russo-Ukrainian Stance} &
  1,460 &
  1,321 \newline (90.48\%) &
  Pro-Ukraine (63.90\%/48.52\%) \newline Pro-Russia (36.10\%/51.48\%) \\ \bottomrule
\end{tabular}%
}
\caption{A summary of selected datasets and ChatGPT's annotation. The column ``ChatGPT annotation'' shows the volume of ChatGPT's responses matching any candidate label of the dataset, with the percentage representing its proportion to dataset size. The column ``Label (Human/ChatGPT)'' details the proportion of each label's size to dataset size, annotated by human or ChatGPT respectively.}
\label{tab:datasets}
\end{table}

{We use seven public social computing datasets related to distinct social problems.}
Our dataset selection strategy is based on the following requirements: 
\one The datasets must be in English to avoid differences in language provision~\cite{sera2002language}, 
\two The datasets must be annotated by human annotators, as we wish to compare the human annotations with ChatGPT.
We list our targeted annotation tasks and datasets below.








\pb{Dataset 1: Vaccine Stance.}
{We select stance detection dataset towards twitter users' attitudes to COVID-19 vaccine. This dataset, denoted as \texttt{Vaccine Stance}~\cite{poddar2022winds}, contains 5,926 tweets related to COVID-19 vaccine and ChatGPT annotates 5,832 of them. 
}


\pb{Dataset 2: COVID-19 Hate Speech.} We select an anti-Asian hate and counterspeech detection dataset called \texttt{COVID-HATE}~\cite{he2021racism}.
We utilize ChatGPT to annotate 2,280 items from a subset released with the human annotations of \texttt{COVID-HATE}.

\pb{Dataset 3: COVID-19 Fake News.} We select a fake news detection dataset denoted as \texttt{COVID-19 Fake News}~\cite{patwa2020fighting}. This dataset contains news-sharing posts related to COVID-19.
We utilize ChatGPT to annotate 10,540 items from \texttt{COVID-19 Fake News}.




\pb{Dataset 4: Social Bot.}
We select a deepfake detection dataset called \texttt{TweepFake}~\cite{fagni2021tweepfake}.
This dataset contains tweets posted by 23 bots and 17 human accounts verified by the dataset's authors. 
We utilize ChatGPT to annotate 16,235 items from \texttt{TweepFake}.

\pb{Dataset 5: Anti-LGBT Cyberbullying.}
{We select a cyberbully detection dataset denoted as \texttt{Anti-LGBT Cyberbullying}. This dataset is derived from a large-scale hate speech dataset~\cite{kennedy2020constructing} and re-annotated for building a binary-classifier to identify anti-LGBT cyberbully.\footnote{\url{https://www.kaggle.com/datasets/kw5454331/anti-lgbt-cyberbully-texts}} 
We utilize ChatGPT to annotate 4,258 items from \texttt{Anti-LGBT Cyberbullying}.}

\pb{Dataset 6: Clickbait Headlines.}
{We select a clickbait detection dataset denoted as \texttt{Clickbait Headlines}~\cite{chakraborty2016stop}. This dataset contains news headlines of Wikinews posts and articles published on well-known clickbait websites (\eg BuzzFeed, Upworthy, ViralNova). 
We utilize ChatGPT to annotate 31,084 items in from \texttt{Clickbait Headlines}.}

\pb{Dataset 7: Russo-Ukrainian Stance.}
We also investigate how ChatGPT performs on timely and recent domains that have arisen since its launch ($23^{th}$ November, 2022).
{Thus, we select a tweets dataset, denoted as \texttt{Russo-Ukrainian Stance}, for stance detection regarding to debates on the Russo-Ukrainian War~\cite{RU-echo}.} 
This dataset collects relevant tweets annotated based on its stance towards the invasion.
We utilize ChatGPT to annotate 1,321 items from \texttt{Russo-Ukrainian Stance}.


\subsection{ChatGPT Annotation}\label{sec:chatgpt_annotation}


For each annotated dataset, we try to recreate the same annotations using ChatGPT.
We utilize OpenAI API,
configured with ChatGPT module \texttt{gpt-3.5-turbo-0631}, to annotate each target dataset.
We rely on an official prompt example
for classification tasks from OpenAI.\footnote{\url{https://platform.openai.com/docs/guides/completion/prompt-design}}

In the official document, most prompts are imperative sentences starting with a verb. 
As such, we choose ``Classify'', which is frequently used in annotation work.
We use this verb to design our prompt. 
According to the official template, we find that starting a new row with a word describing the subject and object in the prompt is effective.
Thus, we follow this pattern, injecting the subject and objective of the annotation task here.
{Another benefit of this template is that it 
is flexible to inject text input to annotate and specify a desired format for ChatGPT to respond with its annotation.}

Based on this template, we modify this example into a generalized prompt template applicable to our seven distinctive datasets.
The template can be adjusted to be applied for annotating text in different context as shown as follows:

\begin{tcolorbox}[
    standard jigsaw,
    opacityback=0
]
\footnotesize
\texttt{Classify the text about \textcolor{blue}{\{Topic\}} with a label from [\textcolor{blue}{Label 1, Label 2, ...}].\newline Text: ``\textcolor{blue}{\{text to classify\}}''.\newline Desired format: <label\_for\_classification>}
\end{tcolorbox}


\noindent
where \texttt{\{Topic\}} refers to the topic or background of the text; The \texttt{[label1, label2, ...]} refers to the set of candidate labels for ChatGPT to annotate the text; and \texttt{\{text to classify\}} refers to the text input for ChatGPT to produce label. {In addition, the plain-language index \texttt{Desired format} indicates that ChatGPT should only respond using the label without any other text.}



We then apply this template to generate a ChatGPT prompt according to the dataset's original annotation strategy. 
For example, the \texttt{COVID-HATE} tweets focus on COVID-19 
and are classified into three labels -- \textit{Neutral}, \textit{Counterspeech}, and \textit{Hate}. Accordingly, we inform ChatGPT of the tweets' topic (as COVID-19) and ask it to classify the tweets into the same three labels. 
Below is an example of such a  prompt, where the bold text refers to the content we edited within the generalized template: 

\begin{tcolorbox}[
    standard jigsaw,
    opacityback=0
]
\footnotesize
\texttt{Classify the text about \textcolor{blue}{COVID-19} with a label from [\textcolor{blue}{Hate, Counterspeech, Neutral}].\newline Text: ``\textcolor{blue}{for the last f**king time.... CORONAVIRUS IS NO EXCUSE TO BE RACIST AGAINST ASIANS https://t.co/nBHTadCKzK}''.\newline Desired format: <label\_for\_classification>}
\end{tcolorbox}

Following this, we pass all data to ChatGPT for annotation. When ChatGPT responds to such a prompt, it is necessary to parse the response and extract its annotation. We consider a response is parsable only it follows the desired format -- only providing a label without any other text. Thus, we extract ChatGPT’s annotation by matching its response to any candidate labels for the dataset.

In all, we only encounter an average of 3.13\% ($SD=2.99\%$) responses per-dataset that fail to be parsed. Note, a small number ($0.1\%$) of failed cases are because ChatGPT states there is not enough information for it to make a decision. For example, a failed response from Russo-Ukrainian Stance dataset states -- ``\textit{Cannot classify the given text about Russo-Ukrainian War with the label [Pro-Russia, Pro-Ukraine] as it does not provide any relevant information about the topic.}''

We emphasize that there are many ways in which our methodology could be expanded and refined. Our future work will involve exploring alternative forms of prompt formulation and response mining.


\section{Results and Analysis}\label{sec:analysis}

To evaluate the performance of ChatGPT, we compare ChatGPT's annotation against the original human annotations contained within each dataset.
We treat the original human annotations as the gold standard that ChatGPT must predict. As such, we treat ChatGPT as a prediction engine, which we can then evaluate using traditional classifier metrics.
We use weighted F1-score to evaluate ChatGPT's performance on data annotation. 
Given a dataset, a higher F1-score indicates that ChatGPT provides annotations that are \emph{more} similar to humans.


\subsection{Annotation Data Summary}

Table~\ref{tab:datasets} presents statistics for the ChatGPT annotation results. The annotated datasets contain 73,493 text items.
ChatGPT annotates 71,484 (97.27\%) items. 
For the remainder, 2\% of responses do not match any candidate label in the dataset, and 0.73\% of responses fail due to API errors.
This confirms that ChatGPT can generate easily extractable annotation labels in desired format in most cases. 

\subsection{Results Summary}
\label{sec:manual_annotation}

\begin{table}[t]
\centering
\resizebox{\columnwidth}{!}{%
\begin{tabular}{|L{12em}|C{6em}C{6em}C{6em}|}
\toprule
\textbf{Dataset} &
  \textbf{w-Recall\newline(Accuracy)} &
  \textbf{w-Precision} &
  \textbf{w-F1-score} \\ \midrule
\textbf{Vaccine Stance} & 59.81\% & 66.11\% & 59.17\% \\ \midrule
\textbf{COVID-19 Hate Speech} & 52.24\% & 55.61\% & 51.88\% \\ \midrule
\textbf{COVID-19 Fake News} & 83.75\% & 85.55\% & 83.43\% \\ \midrule
\textbf{Social Bot} & 64.96\% & 65.33\% & 63.70\% \\ \midrule
\textbf{Anti-LGBT Cyberbullying} & 79.08\% & 87.17\% & 80.03\% \\ \midrule
\textbf{Clickbait Headlines} & 89.66\% & 90.92\% & 89.56\% \\ \midrule
\textbf{Russo-Ukrainian Stance} & 75.85\% & 79.83\% & 76.26\% \\ \bottomrule
\end{tabular}%
}
\caption{A summary of ChatGPT's annotation performance on selected datasets. The prefix \enquote{w-} notes that the measurements' calculations are weighted average by the number of human-annotated items for the label.}
\label{tab:performance}
\end{table}

\begin{figure*}[t]
    \centering
    \begin{subfigure}[b]{0.24\textwidth}
        \centering
        \includegraphics[width=\textwidth]{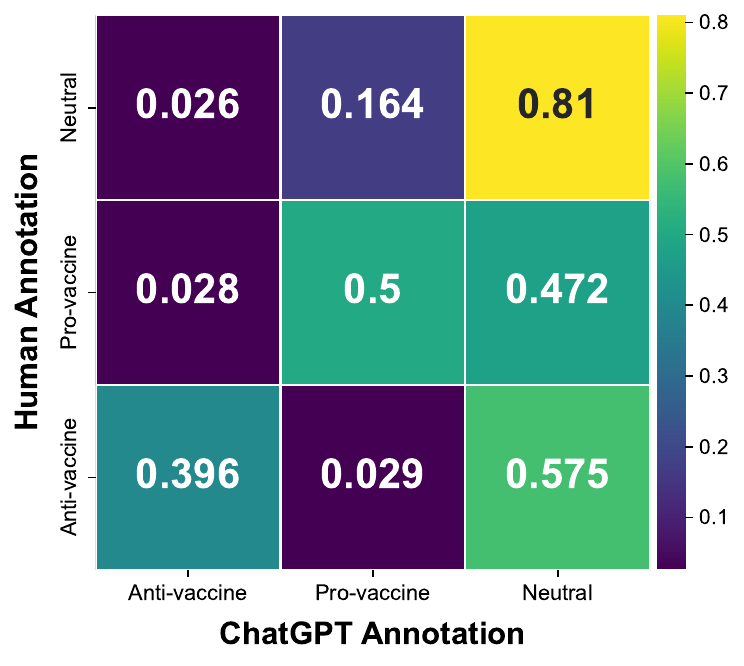}
        \caption{Vaccine Stance}%
        \label{fig:cm_vaccine}
    \end{subfigure}
    \begin{subfigure}[b]{0.24\textwidth}
        \centering
        \includegraphics[width=\textwidth]{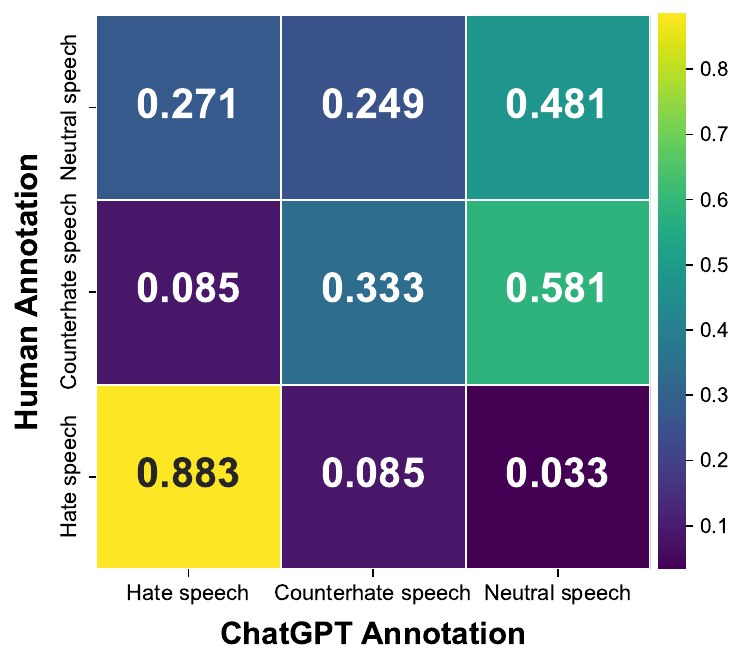}
        \caption{COVID-19 Hate Speech}%
        \label{fig:cm_hate}
    \end{subfigure}
    \begin{subfigure}[b]{0.24\textwidth}
        \centering
        \includegraphics[width=\textwidth]{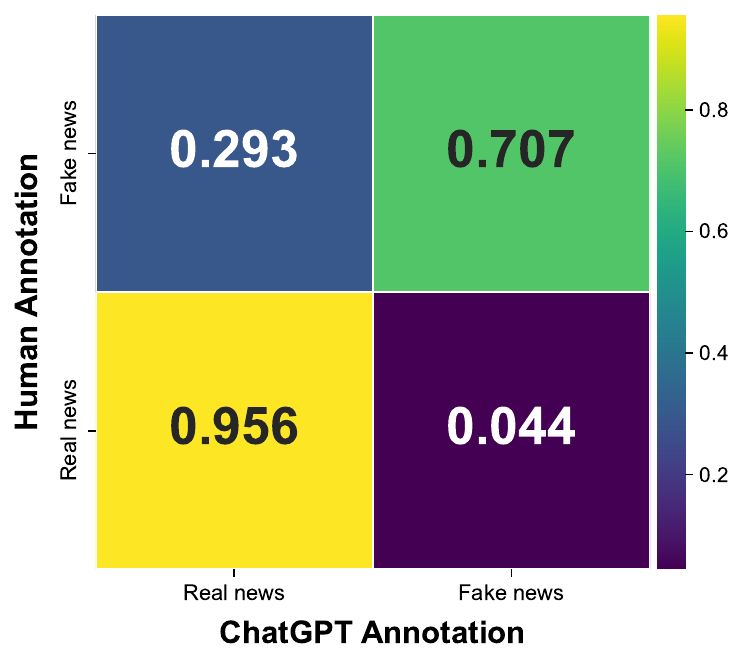}
        \caption{COVID-19 Fake News}%
        \label{fig:cm_fake}
    \end{subfigure}
    \begin{subfigure}[b]{0.24\textwidth}
        \centering
        \includegraphics[width=\textwidth]{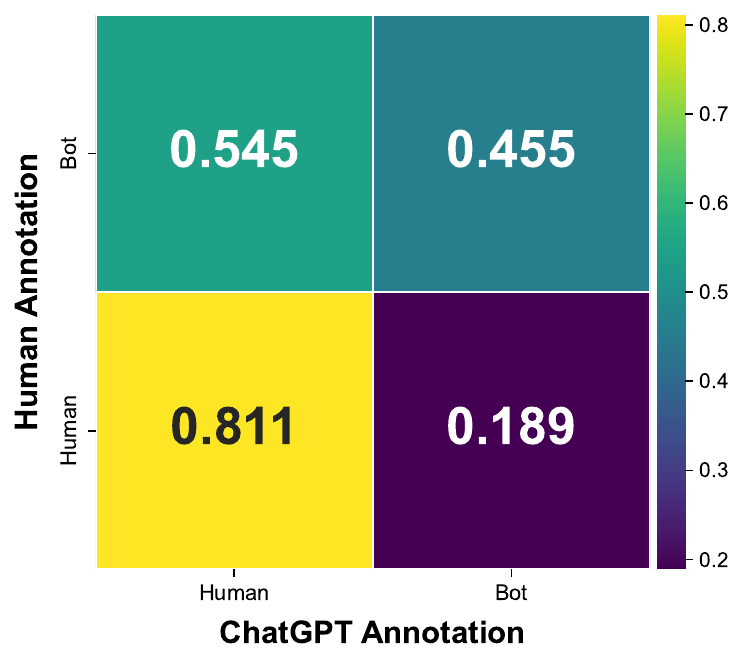}
        \caption{Social Bot}%
        \label{fig:cm_bot}
    \end{subfigure}
    \vskip\baselineskip
    \begin{subfigure}[b]{0.24\textwidth}
        \centering
        \includegraphics[width=\textwidth]{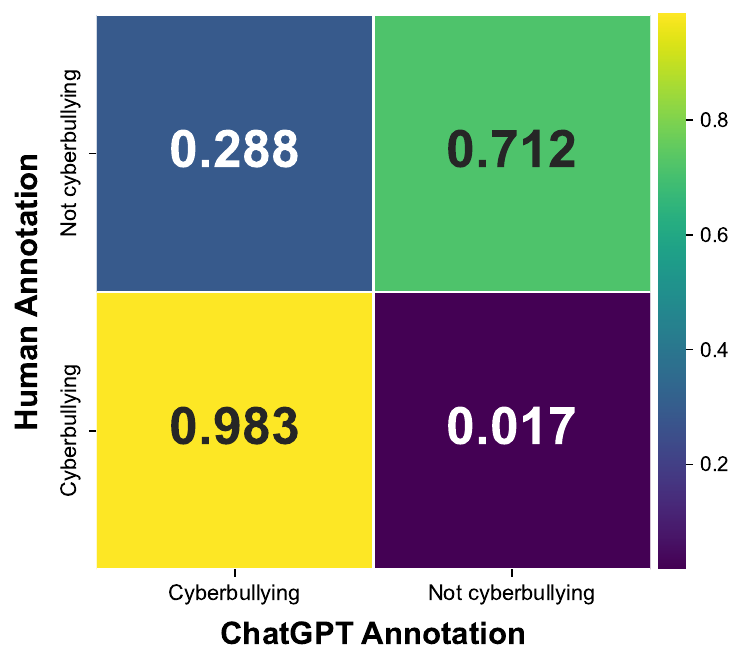}
        \caption{Anti-LGBT Cyberbullying}%
        \label{fig:cm_lgbt}
    \end{subfigure}
    \begin{subfigure}[b]{0.24\textwidth}
        \centering
        \includegraphics[width=\textwidth]{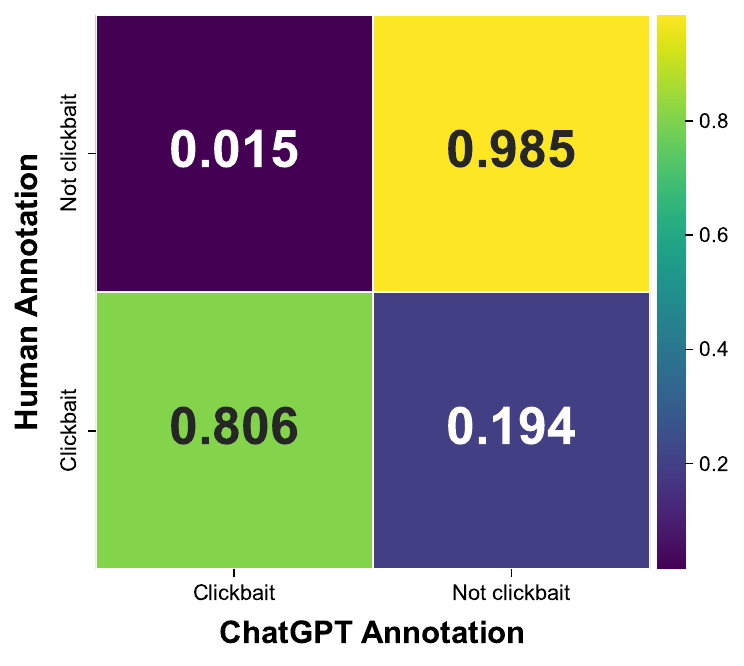}
        \caption{Clickbait Headlines}%
        \label{fig:cm_clickbait}
    \end{subfigure}
    \begin{subfigure}[b]{0.24\textwidth}
        \centering
        \includegraphics[width=\textwidth]{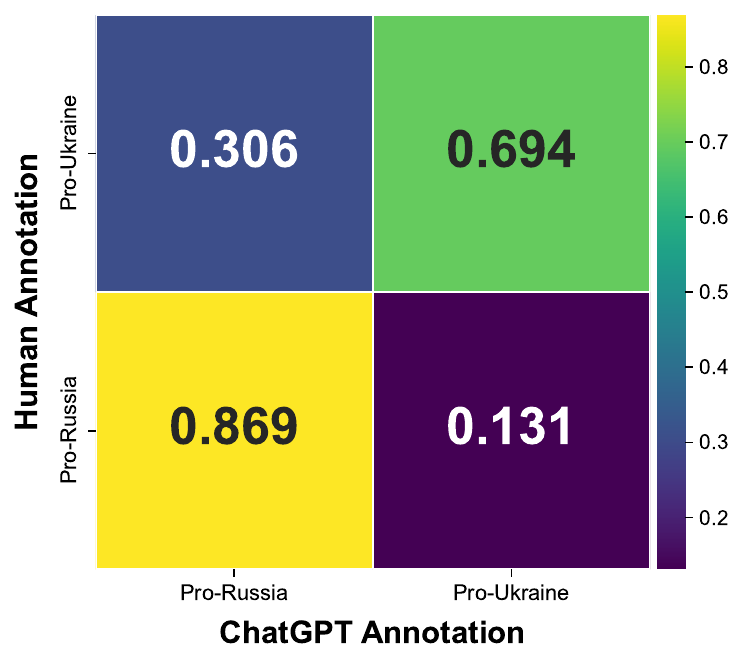}
        \caption{Russo-Ukrainian Stance}%
        \label{fig:cm_RU}
    \end{subfigure}
    \caption{
    The confusion matrices of ChatGPT's annotations for the seven annotation datasets. The y-axis label for a given row shows human label of text, and values in each cell show the percentage of those text annotated in the corresponding x-axis label by ChatGPT.
    }
    \label{fig:cm}
\end{figure*}

\begin{table}[t]
\centering
\resizebox{\columnwidth}{!}{%
\begin{tabular}{|L{12em}|L{8em}C{5em}C{5em}C{5em}|}
\toprule
\textbf{Dataset} & \textbf{Label} & \textbf{Recall} & \textbf{Precision} & \textbf{F1-score} \\ \midrule
\textbf{Vaccine Stance} & Anti-vaccine \newline Pro-vaccine \newline Neutral & 39.65\% \newline 50.02\% \newline 80.99\% & 80.13\% \newline 74.03\% \newline 50.25\% & 53.05\% \newline 59.70\% \newline 62.02\% \\ \midrule
\textbf{COVID-19 Hate Speech} & Hate  \newline Counterhate \newline Neutral & 88.27\% \newline 33.33\% \newline 67.19\% & 48.08\% \newline 31.79\% \newline 31.79\% & 62.25\% \newline 32.54\% \newline 56.03\% \\ \midrule
\textbf{COVID-19 Fake News} & Real news \newline Fake news & 95.63\% \newline 70.70\% & 78.18\% \newline 93.65\% & 86.03\% \newline 80.57\% \\ \midrule
\textbf{Social Bot} & Human \newline Bot & 81.12\% \newline 45.54\% & 64.16\% \newline 66.75\% & 71.65\% \newline 54.14\% \\ \midrule
\textbf{Anti-LGBT Cyberbullying} & Cyberbully \newline Not cyberbully & 98.28\% \newline 71.17\% & 58.43\% \newline 99.02\% & 73.29\% \newline 82.81\% \\ \midrule
\textbf{Clickbait Headlines} & Clickbait \newline Not clickbait  & 80.57\% \newline 98.53\% & 98.17\% \newline 83.86\% & 88.50\% \newline 90.60\% \\ \midrule
\textbf{Russo-Ukrainian Stance} & Pro-Ukraine \newline Pro-Russia & 69.35\% \newline 86.91\% & 90.02\% \newline 62.50\% & 78.34\% \newline 72.71\% \\ \bottomrule
\end{tabular}%
}
\caption{A summary of ChatGPT's label-wise annotation performance on selected datasets.}
\label{tab:label_performance}
\end{table}

Table~\ref{tab:performance} presents the weighted recall, precision, and F1-score for ChatGPT's predictions for each dataset. 
For the seven datasets, ChatGPT achieves an average weighted F1-score of 72.00\% ($SD=13.90\%$).
{This suggests that, \textit{as a data annotator, ChatGPT has the potential to generate some annotations similar to humans.} However, its performance varies across different domains. ChatGPT presents perform well on COVID-19 Fake News, Anti-LGBT Cyberbullying, Clickbait Headlines, and Russo-Ukrainian Stance datasets (weighted F1-score $>75\%$). In contrast, ChatGPT performs poorly on Vaccine Stance, COVID-19 Hate Speech, and Social Bot datasets (weighted F1-score $<65\%$).}

We next explore how ChatGPT performs on different labels in each annotation dataset.
Figure~\ref{fig:cm} presents the confusion matrices for the seven datasets. For each matrix, the y-axis refers to the ground truth human labels, and the x-axis refers to ChatGPT's labels.
The value in a cell, row \textbf{\emph{i}} and column \textbf{\emph{j}}, presents the proportion of text with label \textbf{\emph{i}} that are annotated with label \textbf{\emph{j}} by ChatGPT. For all annotation tasks, the proportions of correctly annotated tweets by ChatGPT vary per distinct label.
For example, Figure~\ref{fig:cm_vaccine} presents the confusion matrix for the Vaccine Stance dataset. Here, we see that ChatGPT corrects labels just 39.6\% of ``Anti-vaccine'' \vs 50\% for ``Pro-vaccine'' \vs 81\% for ``Neutral''. In this case, ChatGPT is significantly more accurate at annotating text expressing neutral stance speech than those expressing anti-vaccine or pro-vaccine stance. {In all, \textit{five out of seven} datasets present such an imbalance, where a gap of more than 25\% exists between labels' accuracy.} These results suggest that \textit{for a given annotation task, ChatGPT's accuracy varies heavily across different labels.}


\subsection{Task Analysis and Implications}

Next, we dive into each annotation task to present implications according to ChatGPT's performance. To assist the following analyses on ChatGPT's label-wise performance, we also present ChatGPT's recall, precision, and F1-score for each annotation label in Table~\ref{tab:label_performance}.

\pb{Rank 1\textsuperscript{st}: Clickbait Headlines.}
For Clickbait Headlines, ChatGPT's overall performance ranks \textit{first} out of seven annotation tasks by a weighted F1-score of 89.56\%.
ChatGPT correctly identifies 89.66\% clickbait or non-clickbait news headlines. This is the highest accuracy among the annotation datasets. In addition, the weighted precision rate of 90.92\% suggests that ChatGPT can effectively annotate clickbait news headlines, while only introducing a small number of false positives. {In addition, ChatGPT attains recall, precision, and F1-score exceeding 80\% across all labels. Specifically, the corresponding confusion matrix (Figure~\ref{fig:cm_clickbait}) shows that ChatGPT can correctly identify 80.6\% of clickbait and 98.5\% of non-clickbait headlines. In conclusion, \textit{ChatGPT shows better overall performance on Clickbait Headlines than the other six annotation datasets.}}

\pb{Rank 2\textsuperscript{nd}: COVID-19 Fake News.} For COVID-19 Fake News, ChatGPT's overall performance ranks \textit{second} out of seven annotation tasks, with a weighted F1-score of 83.43\%.
{ChatGPT correctly distinguish 83.75\% of news as real or fake. However, ChatGPT's performance across different news varies when measured by recall and precision. For real news, ChatGPT achieves a high recall of 95.63\%, but a lower precision of 78.18\%. For fake news, ChatGPT only attains a recall of 70.7\%, but a higher precision of 93.65\%. Such a pattern is caused by the existence of many false positives of real news annotated by ChatGPT. Moreover, corresponding confusion matrix (Figure~\ref{fig:cm_fake}) shows that, while ChatGPT can identify 95.6\% real news, 29.3\% fake news are misidentified as real. \textit{These results suggest that, for COVID-19 Fake News, ChatGPT is able to annotate real news with high accuracy, but also has a tendency to misidentify fake news as real.} As fake news detection mainly relies on news content when training models~\cite{shu2017fake, zhou2020survey}, we conjecture \textit{such a limitation is caused by a lack of comprehensive COVID-19 fake news data for training ChatGPT}.}

\pb{Rank 3\textsuperscript{rd}: Anti-LGBT Cyberbullying.} 
For Anti-LGBT Cyberbullying annotation, ChatGPT's overall performance ranks \textit{third} out of seven tasks, with a weighted F1-score of 80.03\%. 
{Overall, ChatGPT correctly annotates 79.08\% posts as cyberbully or not. However, when annotating cyberbully posts, ChatGPT achieves 98.28\% recall, but only attain 58.43\% precision. Such a pattern indicates the existence of false positives among cyberbully posts annotated by ChatGPT. The confusion matrix supports this by showing that ChatGPT only correctly annotates 71.2\% non-cyberbully posts and 28.8\% non-cyberbully posts are misidentified as cyberbully. To summarize, \textit{ChatGPT is better at annotating cyberbully posts (using our prompt) when compared against other posts containing non-cyberbully expressions. However, ChatGPT often overestimates non-cyberbully posts as cyberbully}.}



\pb{Rank 4\textsuperscript{th}: Russo-Ukrainian Stance.}
For the Russo-Ukrainian Stance detection dataset, ChatGPT's overall performance ranks \textit{fourth} out of seven tasks, with a weighted F1-score of 76.26\%.
Note, the invasion of Ukraine took place after the training date of our ChatGPT version.
Overall, ChatGPT correctly annotates 75.85\% of tweets' stance. We find that, for the pro-Ukraine tweets, ChatGPT reports a high precision of 90.02\%, with a low recall of 69.35\%. In contrast, for the pro-Russia tweets, ChatGPT reports a high recall of 86.91\%, but with a low precision of 62.50\%. This means that, when ChatGPT annotates pro-Ukraine tweets, it is usually correct (high precision), but the same is not true for annotating a tweet pro-Russia (low precision). The corresponding confusion matrix (Figure~\ref{fig:cm_RU}) shows that 30.6\% pro-Ukraine tweets are mis-annotated as pro-Russia by ChatGPT. 
In summary, \textit{when ChatGPT labels a tweet as expressing a pro-Ukraine stance, it is usually correct}.
Yet, this comes at the cost of a low recall rate.
In contrast, \textit{ChatGPT has more false positives when labeling tweets as pro-Russian, but does have a higher recall.}

\pb{Rank 5\textsuperscript{th}: Social Bots.}
For the Social Bot dataset, ChatGPT's overall annotation performance ranks \textit{fifth} out of seven annotation datasets, with a weighted F1-score of 63.70\%. 
Overall, ChatGPT correctly annotates 64.96\% of tweets as posted by humans or bots. {For precision, ChatGPT only attains 64.16\% for human tweets and 66.75\% for bot tweets. This suggests ChatGPT's ability to reproduce precise annotations is still limited on social bot detection, i.e., ChatGPT often reports false positives for both human and bot tweets. 
For recall, ChatGPT attains 81.12\% for human tweets but only 45.54\% for bot tweets. Such a significant difference implies that, while ChatGPT manages to identify most human tweets, it often fails to distinguish bot tweets from human ones. The corresponding confusion matrix supports this by showing that 54.5\% of bot tweets are mis-annotated, compared to only 18.9\% human tweets. In summary, \textit{ChatGPT has the potential to act as an annotator to distinguish most human content from bot content.
Yet, ChatGPT's capability is limited when identifying content generated by social bots.
In our case, ChatGPT often accidentally annotates social bot tweets as human-generated. As a result, its annotation for human tweets can involves lots of false positives}.

\pb{Rank 6\textsuperscript{th}: Vaccine Stance.}
For the Vaccine Stance dataset, ChatGPT's overall performance ranks \textit{sixth} out of seven datasets, with a weighted F1-score of 59.17\%. Overall, ChatGPT correctly annotates 59.81\% of tweets' stance towards COVID-19 vaccine. 
ChatGPT is more precise when detecting tweets expressing anti-vaccine (precision $=80.13\%$) and pro-vaccine stance (precision $=74.03\%$), compared to tweets with neutral expression (precision $=50.25\%$). In the meantime, ChatGPT's recall varies across these three labels. While ChatGPT achieves a high recall of 80.99\% for neutral tweets, it attains only 39.65\% for anti-vaccine tweets and 50.02\% for pro-vaccine tweets. 
This indicates that ChatGPT is conservative when annotating tweets' stance towards COVID-19 vaccine and often mis-labels tweets' stance as neutral.
According to the corresponding confusion matrix (Figure~\ref{fig:cm_vaccine}), ChatGPT mis-annotates 57.5\% of anti-vaccine tweets and 47.2\% pro-vaccine tweets as neutral tweets. 
Therefore, \textit{ChatGPT performs poorly in labeling tweets' stance on the COVID-19 vaccine. This is because many anti-vaccine or pro-vaccine tweets are mislabeled as neutral by ChatGPT.}

\pb{Rank 7\textsuperscript{th}: COVID-19 Hate Speech.}
For the COVID-19 Hate Speech dataset, ChatGPT's performance ranks \textit{seventh} out of seven, with a weighted F1-score of 51.88\%. 
Overall, ChatGPT correctly annotates 52.24\% of tweets as expressing or countering anti-Asia hate. 
Specifically, for hate speech, ChatGPT attains a F1-score of 62.25\% with a high recall of 88.27\%. For counterhate speech, ChatGPT attains a very low F1-score of 32.54\%, with a recall of 33.33\%. This suggests that ChatGPT's performance varies when annotating hate and counterhate content on COVID-19. Importantly, ChatGPT's precision for all three labels is lower than 50\%, confirming its annotation for each label involves many false positives. As highlighted by the corresponding confusion matrix (Figure~\ref{fig:cm_hate}), ChatGPT mis-labels 27.1\% of neutral tweets as hate tweets, and 24.9\% neutral tweets as counterhate tweets. Meanwhile, it mis-annotates 58.1\% of counterhate tweets as neutral. In summary, \textit{ChatGPT's annotations are inaccurate for the COVID-19 Hate Speech task. ChatGPT often mis-annotates neutral content as hate speech, and fails to distinguish counterhate speech from neutral content.}

\section{GPT-Rater: Predicting ChatGPT Suitability}

\begin{table*}[t]
\centering
\resizebox{\textwidth}{!}{%
\begin{tabular}{|L{12em}|C{5em}C{5em}C{5em}C{5em}C{5em}C{5em}C{5em}C{5em}C{5em}C{5em}|}
\toprule
& \multicolumn{2}{c}{\textbf{SVM}} & \multicolumn{2}{c}{\textbf{Random Forest}} & \multicolumn{2}{c}{\textbf{Logistic regression}} & \multicolumn{2}{c}{\textbf{XGBoost}} & \multicolumn{2}{c|}{\textbf{LightGBM}} \\ 
\cmidrule(lr){2-3}\cmidrule(lr){4-5}\cmidrule(lr){6-7}\cmidrule(lr){8-9}\cmidrule(lr){10-11}
& \textbf{Accuracy} & \textbf{F1-score} & \textbf{Accuracy} & \textbf{F1-score} & \textbf{Accuracy} & \textbf{F1-score} & \textbf{Accuracy} & \textbf{F1-score} & \textbf{Accuracy} & \textbf{F1-score}\\ \midrule
\textbf{Vaccine Stance} & 63.32\%\newline(0.88\%) & 74.37\%\newline(0.77\%) & 61.55\%\newline(0.94\%) & 74.02\%\newline(0.66\%) & 62.94\%\newline(0.88\%) & 74.14\%\newline(0.73\%) & 61.27\%\newline(1.16\%) & 70.69\%\newline(0.98\%) & 62.17\%\newline(1.34\%) & 71.98\%\newline(1.11\%) \\ \midrule
\textbf{COVID-19 Hate Speech} & 68.55\%\newline(1.92\%) & 70.51\%\newline(1.92\%) & 66.63\%\newline(2.09\%) & 70.23\%\newline(1.90\%) & 67.45\%\newline(2.12\%) & 70.21\%\newline(2.01\%) & 67.52\%\newline(2.24\%) & 69.42\%\newline(2.08\%) & 68.06\%\newline(1.98\%) & 70.11\%\newline(1.92\%) \\ \midrule
\textbf{COVID-19 Fake News} & 85.16\%\newline(0.24\%) & 91.84\%\newline(0.12\%) & 84.24\%\newline(0.70\%) & 91.40\%\newline(0.14\%) & 84.36\%\newline(0.23\%) & 91.41\%\newline(0.12\%) & 86.38\%\newline(0.40\%) & 92.34\%\newline(0.22\%) & 86.17\%\newline(0.40\%) & 92.27\%\newline(0.22\%) \\ \midrule
\textbf{Social Bot} & 74.39\%\newline(0.59\%) & 81.95\%\newline(0.45\%) & 68.93\%\newline(0.48\%) & 79.98\%\newline(0.30\%) & 72.19\%\newline(0.57\%) & 80.31\%\newline(0.44\%) & 71.10\%\newline(0.61\%) & 79.11\%\newline(0.45\%) & 71.90\%\newline(0.64\%) & 80.03\%\newline(0.47\%) \\ \midrule
\textbf{Anti-LGBT Cyberbullying} & 81.05\%\newline(0.61\%) & 89.12\%\newline(0.33\%) & 79.28\%\newline(0.22\%) & 88.41\%\newline(0.12\%) & 80.40\%\newline(0.51\%) & 88.83\%\newline(0.28\%) & 81.55\%\newline(0.93\%) & 89.11\%\newline(0.54\%) & 81.61\%\newline(0.71\%) & 89.23\%\newline(0.42\%) \\ \midrule
\textbf{Clickbait Headlines} & 90.65\%\newline(0.12\%) & 95.04\%\newline(0.06\%) & 89.68\%\newline(0.02\%) & 94.56\%\newline(0.01\%) & 90.72\%\newline(0.15\%) & 95.06\%\newline(0.08\%) & 90.97\%\newline(0.19\%) & 95.15\%\newline(0.10\%) & 90.94\%\newline(0.18\%) & 95.15\%\newline(0.10\%) \\ \midrule
\textbf{Russo-Ukrainian Stance} & 75.97\%\newline(0.20\%) & 86.33\%\newline(0.10\%) & 75.85\%\newline(0.96\%) & 86.02\%\newline(0.56\%) & 76.01\%\newline(0.25\%) & 86.34\%\newline(0.13\%) & 75.56\%\newline(1.83\%) & 85.28\%\newline(1.16\%) & 75.98\%\newline(1.41\%) & 85.75\%\newline(0.88\%) \\
\midrule \midrule 
\textbf{Rank correlation} & 0.964*** & 1.000*** & 0.964*** & 1.000*** & 0.964*** & 1.000*** & 0.964*** & 1.000*** & 0.964*** & 1.000***\\ \bottomrule
\end{tabular}%
}
\caption{A summary of experimented classifiers' average accuracy and F1-score to predict whether ChatGPT will reproduce labels similar to human annotators on seven datasets. The percentage in brackets represents corresponding standard deviation of the average accuracy/F1-score. The row ``Rank correlation'' shows Spearman correlation coefficient with associated statistical significance between accuracy/F1-scores for ChatGPT vs. corresponding GPT-Rater classifier on seven datasets. *** denotes p-value < 0.001.}
\label{tab:classifer_performance}
\end{table*}

The previous section has revealed that ChatGPT's annotation performance varies on a per-task and per-label basis. For it to become an applicable tool, it is therefore necessary to formulate mechanisms to predict for which tasks and data items ChatGPT will perform well. To this end, we propose a tool, \textit{GPT-Rater}, that can estimate how likely ChatGPT will give the correct annotation for a data item. This can help inform researchers on ChatGPT's suitability for their task.

\subsection{GPT-Rater Implementation}

GPT-Rater performs a binary classification task. Its goal is to predict if ChatGPT will be able to correctly select the label for a given task and data item. 
The input feature set is the document embedding of the data item, and the prediction target is a binary label: 1 indicating that ChatGPT will correctly allocate the same label as the human annotator's, and 0 indicating otherwise. For the document embedding, we employ OpenAI's \texttt{text-embedding-ada-002}~\cite{neelakantan2022text} model. 

\pb{Classifiers and parameters.}
We have experimented with five common classification algorithms for GTP-Rater. We summarize their corresponding implementation below:
\begin{itemize}[leftmargin=*]
    \item\textbf{Support vector machine (SVM)}: We implement a SVM classifier using the \texttt{cuml} python package, with parameter setting (\textit{kernel} = ``rbf'', \textit{gamma} = ``scale'', \textit{C} = 1, \textit{random\_state} = 42).
    
    \item\textbf{Random forest}: We implement a random forest classifier using the \texttt{cuml} python package, with parameter setting (\textit{split\_criterion} = ``gini'', \textit{n\_streams} = 1, \textit{random\_state} = 42).
    
    \item\textbf{Logistic regression}: We implement a logistic regression classifier using the \texttt{cuml} python package, with default parameter setting.
    
    \item\textbf{Extreme Gradient Boosting (XGBoost)}: We implement a XGBoost classifier using the \texttt{xgboost} python package, with parameter setting (\textit{tree\_method} = ``gpu\_hist'', \textit{predictor} = ``gpu\_predictor'', \textit{objective} = ``binary:logistic'', \textit{random\_state} = 42).
    
    \item\textbf{Light gradient-boosting machine (LightGBM)}: We implement a LightGBM classifier using the \texttt{lightgbm} python package, with parameter setting (\textit{boosting\_type} = ``gbdt'', \textit{device} = ``gpu'', \textit{random\_state} = 42).
\end{itemize}

\pb{Training and testing.} 
A single GPT-Rater model is trained for each dataset. 
To underpin this, GPT-Rater assumes that the researchers can annotate an initial small number of posts to train it.
Each model is then used to predict if an unseen item from that dataset will be correctly labeled by ChatGPT. A researcher can then use these predictions to decide if they should use ChatGPT. For example, if GPT-Rater estimates that only 10\% of posts will be correctly labelled by ChatGPT, then a researcher should avoid its use.
To measure the number of annotations required this, we later experiment with a range of labeled set sizes. In each case, we split each set into training and testing subsets using an 80:20 ratio. This random splitting and testing process is repeated 100 times to assess the overall performance of our five classifiers. 
Note, the trained GPT-Rater models can then be publicly shared with the community.


\subsection{GPT-Rater Performance}

\begin{figure*}[t]
    \centering
    \begin{subfigure}[b]{0.24\textwidth}
        \centering
        \includegraphics[width=\textwidth]{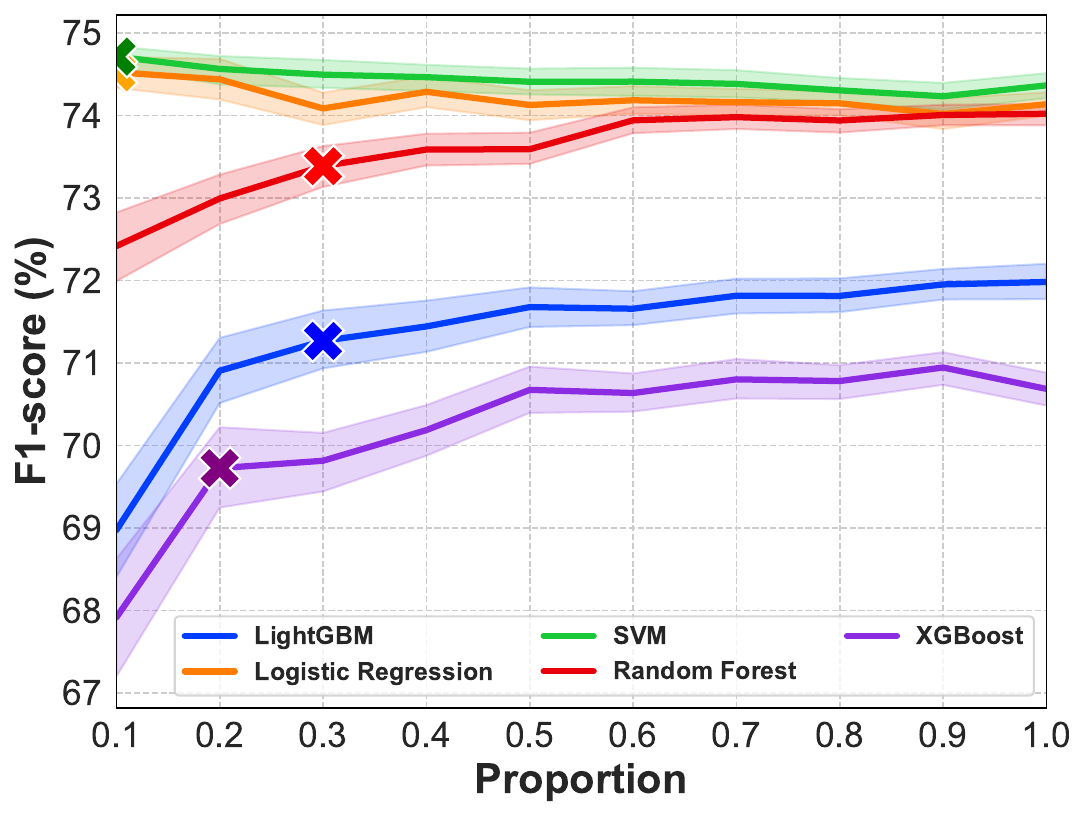}
        \caption{Vaccine Stance}%
        \label{fig:proportion_vaccine}
    \end{subfigure}
    \begin{subfigure}[b]{0.24\textwidth}
        \centering
        \includegraphics[width=\textwidth]{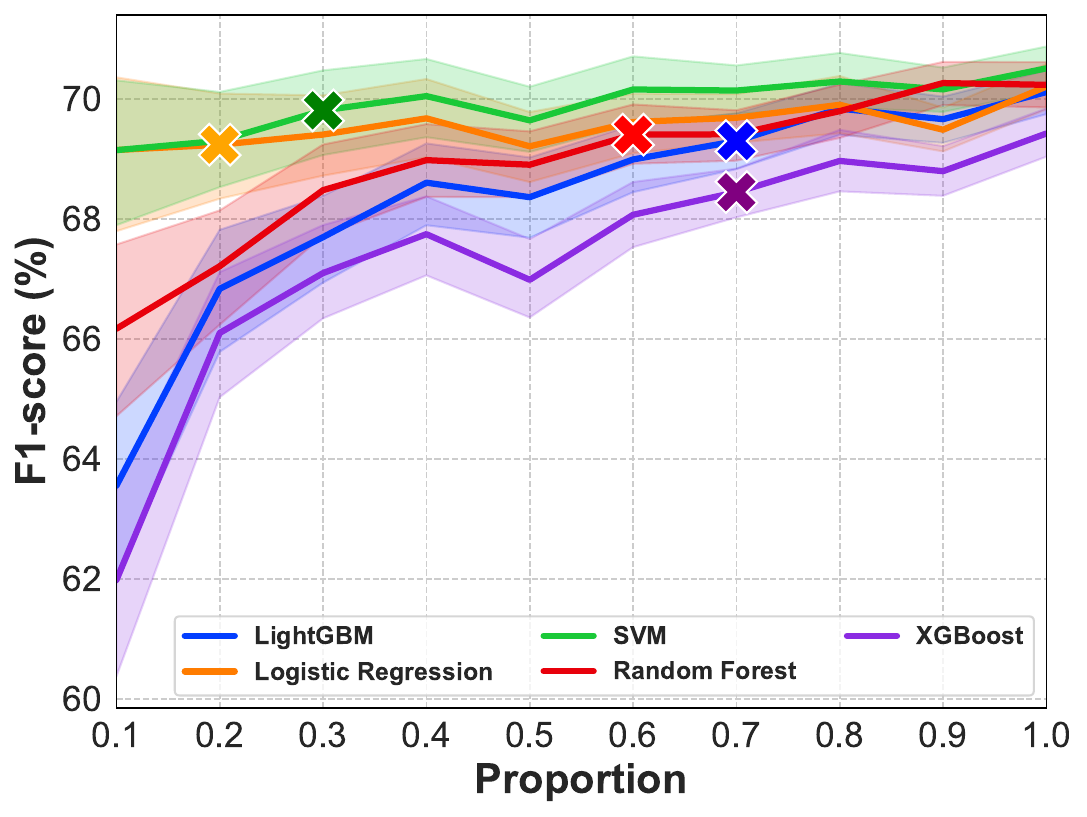}
        \caption{COVID-19 Hate Speech}%
        \label{fig:proportion_hate}
    \end{subfigure}
    \begin{subfigure}[b]{0.24\textwidth}
        \centering
        \includegraphics[width=\textwidth]{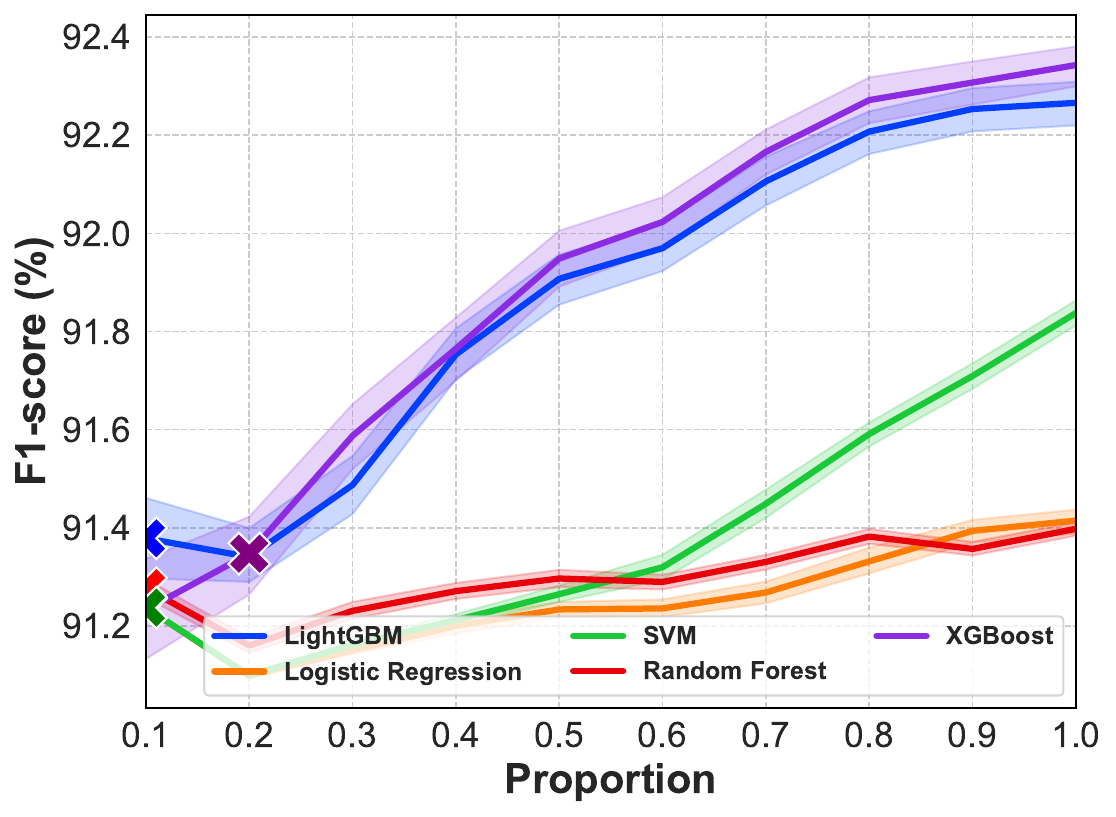}
        \caption{COVID-19 Fake News}%
        \label{fig:proportion_fake}
    \end{subfigure}
    \begin{subfigure}[b]{0.24\textwidth}
        \centering
        \includegraphics[width=\textwidth]{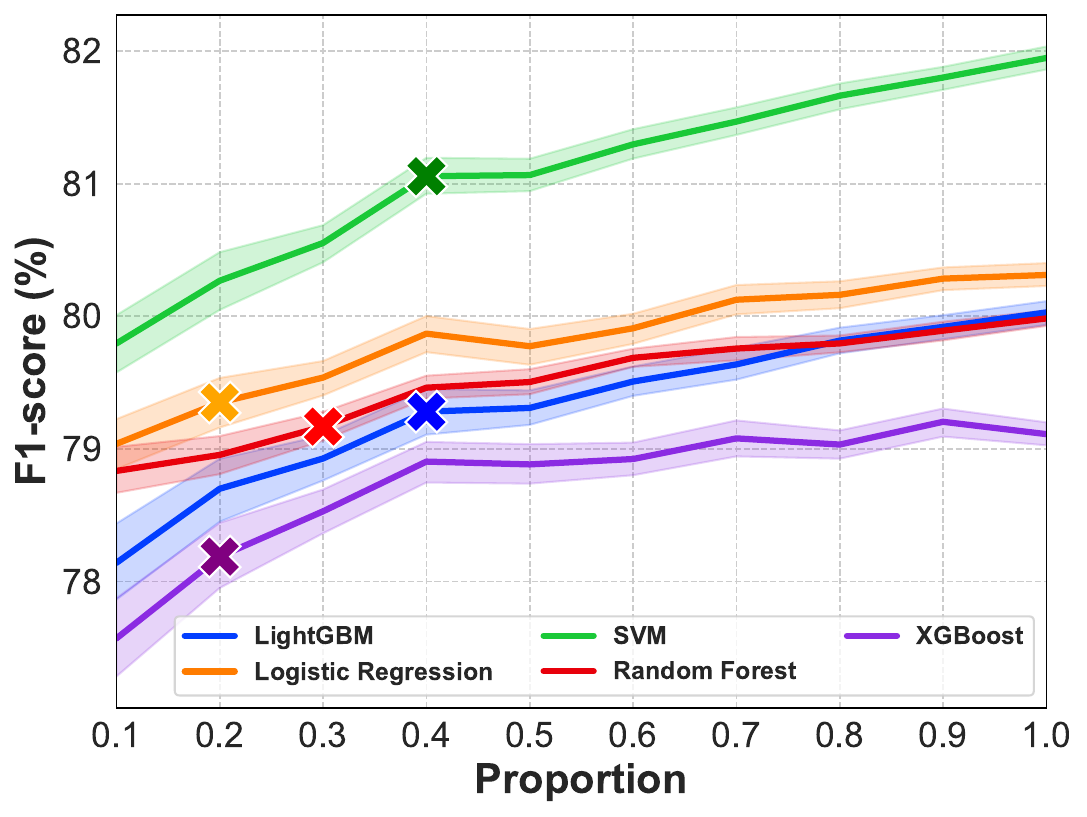}
        \caption{Social Bot}%
        \label{fig:proportion_bot}
    \end{subfigure}
    \vskip\baselineskip
    \begin{subfigure}[b]{0.24\textwidth}
        \centering
        \includegraphics[width=\textwidth]{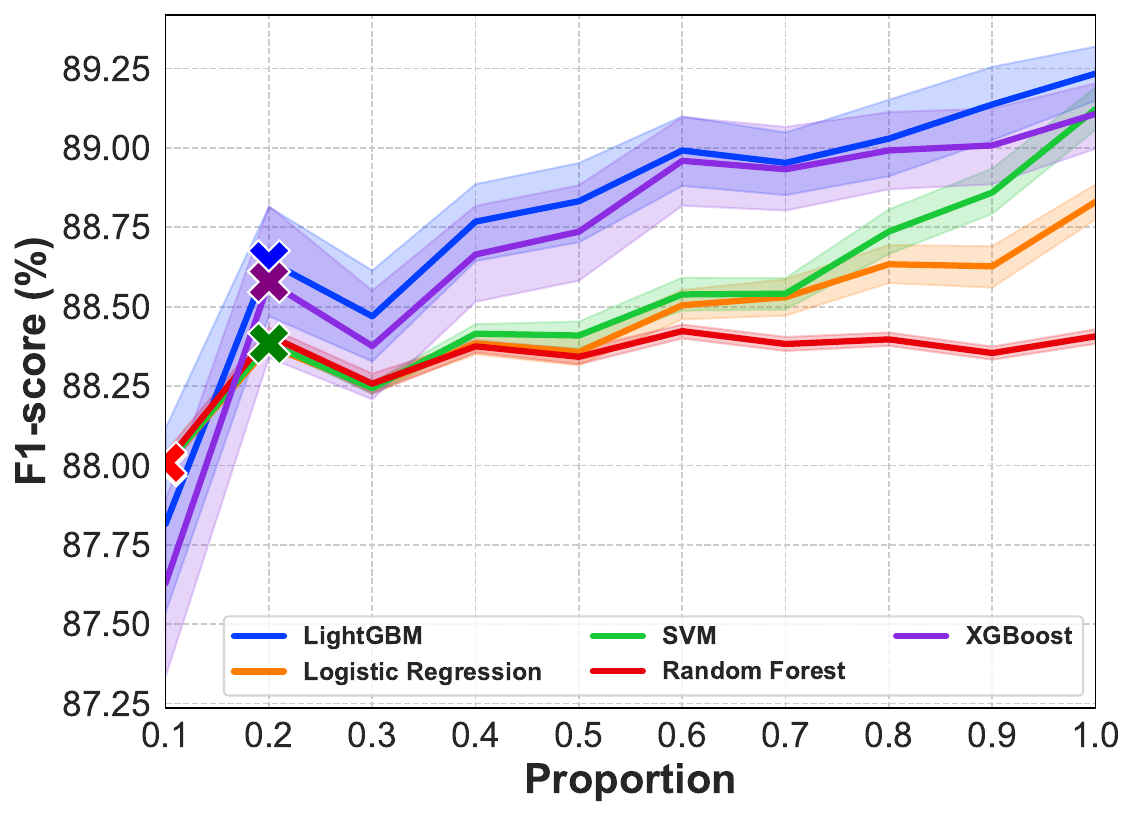}
        \caption{Anti-LGBT Cyberbullying}%
        \label{fig:proportion_lgbt}
    \end{subfigure}
    \begin{subfigure}[b]{0.24\textwidth}
        \centering
        \includegraphics[width=\textwidth]{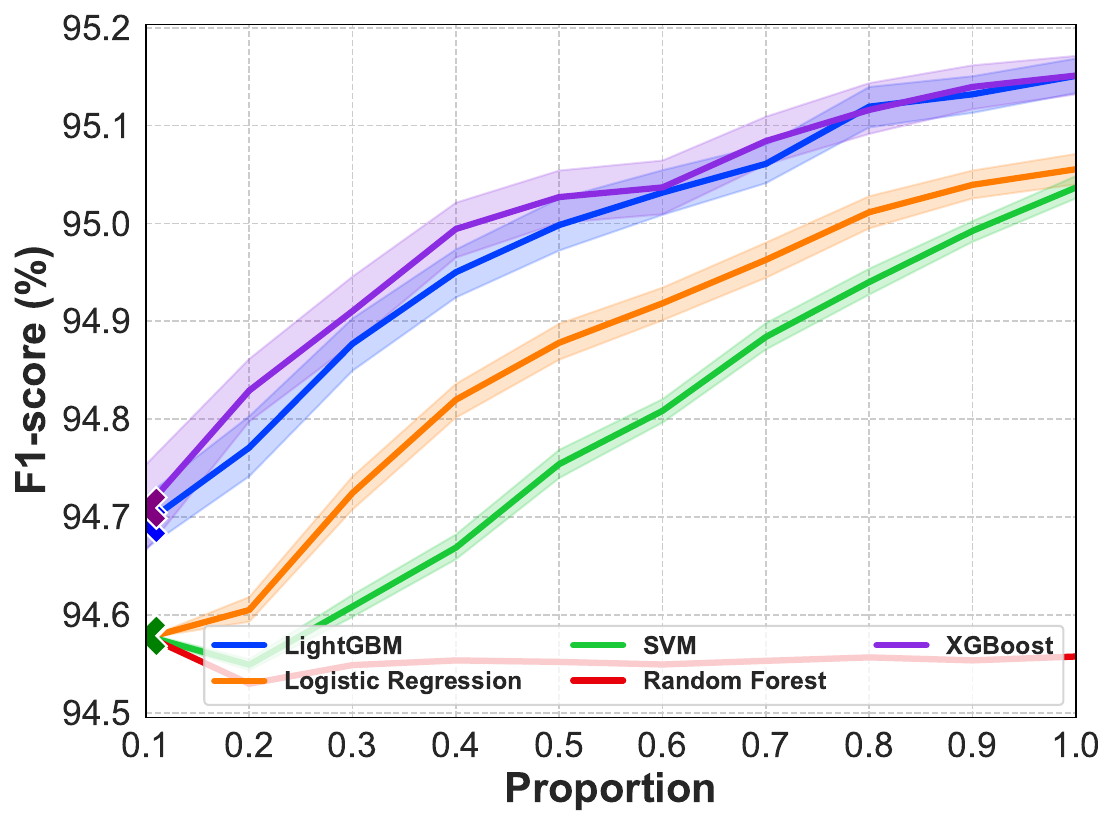}
        \caption{Clickbait Headlines}%
        \label{fig:proportion_clickbait}
    \end{subfigure}
    \begin{subfigure}[b]{0.24\textwidth}
        \centering
        \includegraphics[width=\textwidth]{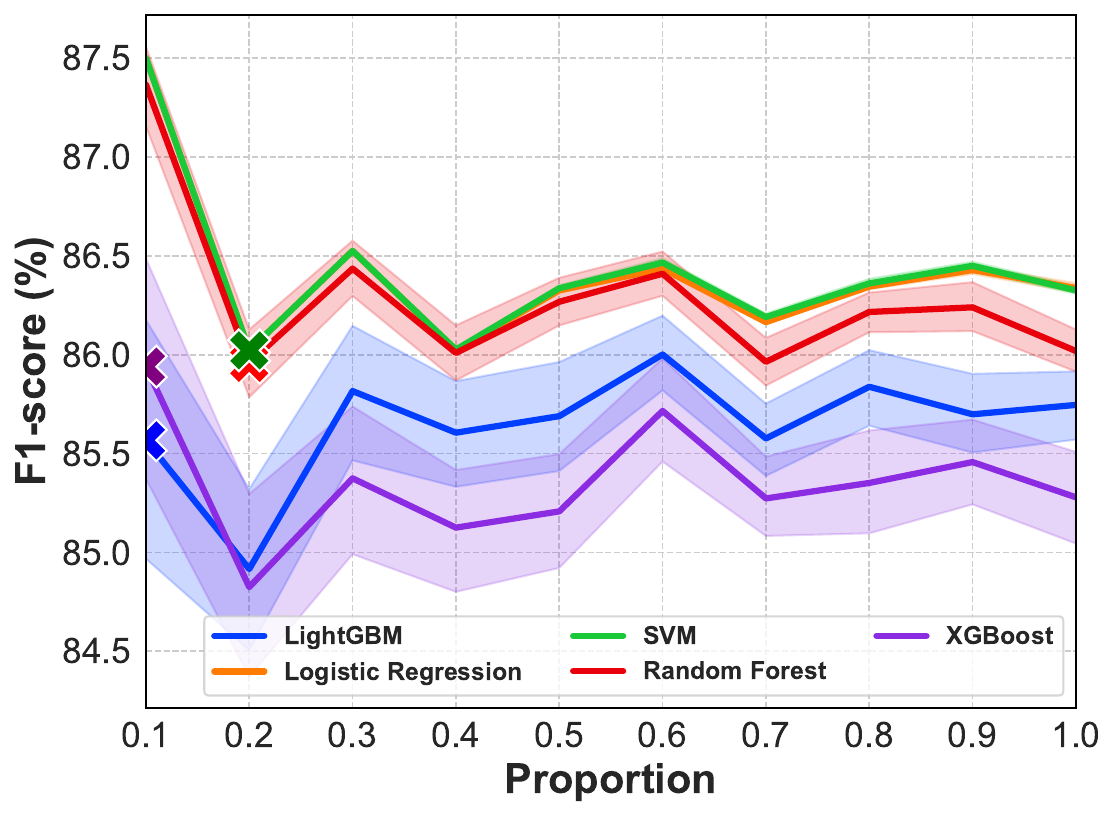}
        \caption{Russo-Ukrainian Stance}%
        \label{fig:proportion_RU}
    \end{subfigure}
    \caption{
    Distribution of F1-scores of GPT-Rater classifiers over subsets by different proportions. {The x-axis ``Proportion'' denotes the proportion in which subsets are sampled from corresponding full datasets to retrain and test GPT-Rater classifiers, within a 80:20 train-test split.} A ``\faRemove'' denotes the lowest proportion where a GPT-Rater classifier can achieve a F1-score close to its F1-score in corresponding full datasets (gap $<1\%$).}
    \label{fig:proportion}
\end{figure*}


\pb{Overall performance.} 
Table~\ref{tab:classifer_performance} summarizes the average and standard deviation of accuracy and F1-score of GPT-Rater across all classifiers on our seven datasets. As an initial step, each dataset involves training GPT-Rater on 80\% of annotated data and testing their performance on the remaining 20\%.
{Our results shows that GPT-Rater predicts ChatGPT's annotation performance well. 
GPT-Rater performs best on the Clickbait Headlines dataset by consistently achieving very high accuracy ($\mu=90.59\%, \sigma=0.13\%$) and F1-scores ($\mu=95.00\%, \sigma=0.05\%$).}
Overall, all classifiers consistently achieve high average F1-scores exceeding 75\% on five out of seven datasets, with small standard deviations below 3\%.
We also note that the ranking of GPT-Rater's average accuracy and F1-score aligns closely with the ranking of ChatGPT's annotation performance across all seven datasets. 
A Spearman correlation analysis reports a significant ($p<0.001$) positive rank correlation between accuracy/F1-score for ChatGPT vs. GPT-Rater on the seven datasets (accuracy: $r=0.964$, F1-score: $r=1.000$) (see Table~\ref{tab:classifer_performance})}
For example, ChatGPT annotations achieve the highest weighted F1-score (89.56\%) for the Clickbait Headlines dataset, while GPT-Rater also gains its best performance for this dataset.
In contrast, in domains where ChatGPT's performance is less robust (such as Vaccine Stance and COVID-19 Hate Speech), all five GPT-Rater classifiers experience a performance drop, with average F1-scores around 70\% for all datasets. This means that GPT-Rater's performance seems reliable only in domains where ChatGPT also performs better. 

To summarize, GPT-Rater shows that ChatGPT's annotation performance is \textit{predictable}, and this predictability remains \textit{consistent}. 
Furthermore, predictions are more successful in domains where ChatGPT excels.
We argue that GPT-Rater can be useful for researchers wishing to estimate how suitable ChatGPT would be for performing their annotation task.

\pb{Performance on subsets.} 
A key limitation of GPT-Rater is that it must be trained for each task and dataset individually. This means that a researcher working on a new task will need to re-train GPT-Rater from scratch.
To evaluate the overhead of this, we next inspect what size of training data is required for GPT-Rater to become usable.
For this, we randomly sampled subsets of our seven dataset respectively in proportions from [0.1, 0.2, 0.3, ..., 1.0]. We then split these subsets by a ratio of 80:20 to retrain and test GPT-Rater classifiers. We repeat the sampling and prediction for 100 times to assess classifiers' overall performance on these subsets.


Figure~\ref{fig:proportion} presents distributions of F1-scores of GPT-Rater classifiers under different training size settings.
The ``\faRemove'' mark represent the location where a GPT-Rater classifier can achieve a F1-score within 1\% of the F1-score in the corresponding full dataset (trained on 80\% of the full annotated data).
We observe that GPT-Rater still attains good performance even with a small proportion of annotated data.
For the four datasets where ChatGPT annotates accurately (F1-score > 75\%), GPT-Rater attains good F1-scores on less than 20\% of labeled data.
However, on the three datasets where ChatGPT gains an F1-score below 65\%, GPT-Rater requires more annotated data by the researcher to attain good F1-scores. For instance, on the COVID-19 Hate Speech task, GPT-Rater requires over 56\% of data to labeled by the researcher to attain a F1-score within 1\% of the F1-score it attains when trained on 80\% of annotated data. This means that it is inappropriate for such tasks.
This shows that GPT-Rater is robust when trained on a subset of annotated data for certain tasks.
In domains where ChatGPT excels, GPT-Rater only requires a small proportion of annotated data by the researcher to perform good prediction for ChatGPT's annotation performance.




\section{Conclusion and Discussion}
\label{sec:discussion}

\pb{Summary.}
This study has investigated ChatGPT's potential to act as a data annotator for social computing data, related to pressing social issues. We have compared ChatGPT's performance on seven datasets against the original human annotations.  
ChatGPT is able to reproduce human labels, achieving an average F1-score of 72.00\%. That said, we discover significant variations in its performance across different domains. This raise the needs for methodologies to identify for which data item ChatGPT will perform well. For this, we propose GPT-Rater to estimate the likelihood that ChatGPT will give the correct annotation for a given data item. Our results show that GPT-Rater attains high average F1-scores, exceeding 75\% with small standard deviations below 3\%.
This suggests that ChatGPT’s annotation performance can be consistently predicted. Additionally, we show that GPT-Rater is robust when trained on a small subset of data annotated by the researcher. This implies its potential to help researchers to preview ChatGPT's annotation capability with far fewer
human labels.

\pb{Implication.}
GPT-Rater can predict ChatGPT's annotation using a small subset of data items labeled by the researcher.
Yet the amount of human labels needed may vary depending on the domain. Our analysis shows that GPT-Rater's prediction is precise, and ChatGPT performs well when GPT-Rater achieves a high F1-score.
From our experiences, we advice researchers to only use GPT-Rater's predictions when the F1-score is above 75\%. If this is not reached, a researcher can either label more data to train GPT-Rater or, alternatively, use more domain-specific knowledge.
After this, a researchers can select a personal threshold for how accurate it needs the ChatGPT annotations to be. For example, if GPT-Rater predicts that ChatGPT will only correctly annotate 20\% of the data, then it may be better to use human annotators.





\pb{Limitations \& Future work.}
{We note that our work has a number of limitations, which form the basis of our future work. 
\textit{First}, this study only examines ChatGPT's annotation performance with a small number of datasets covering five social issues. We would like to further inspect ChatGPT's potential as a annotator on other domains (\eg named-entity recognition~\cite{nadeau2007survey} and topic categorization~\cite{lee2011twitter}), or on other social problems (\eg sexual harassment~\cite{barak2005sexual} and political polarization~\cite{ul2020enemy}). 
\textit{Second}, we acknowledge that we only use a single prompt for annotation, and there are many variants that could be experimented with.
Prompt design is a major theme of future work, which we believe we yield better results.
We are keen to explore how state-of-art prompt-tuning methods, like few-shot prompting~\cite{brown2020language} and Chain-of-Thought technique~\cite{wei2022chain}, can assist with prompt formulation.
\textit{Third}, the design of GPT-Rater is simple, only considering single text embedding as input and five machine learning algorithms. We will fine-tune GPT-Rater to achieve better performance by expanding its input. We further wish to reduce the need for researchers to manually label subsets of data to train GPT-Rater. We hope that our work can act as a catalyst for further work in the area of automated text annotation.}

\bibliographystyle{ACM-Reference-Format}
\bibliography{sample-base}


\begin{thebibliography}{47}


\ifx \showCODEN    \undefined \def \showCODEN     #1{\unskip}     \fi
\ifx \showDOI      \undefined \def \showDOI       #1{#1}\fi
\ifx \showISBNx    \undefined \def \showISBNx     #1{\unskip}     \fi
\ifx \showISBNxiii \undefined \def \showISBNxiii  #1{\unskip}     \fi
\ifx \showISSN     \undefined \def \showISSN      #1{\unskip}     \fi
\ifx \showLCCN     \undefined \def \showLCCN      #1{\unskip}     \fi
\ifx \shownote     \undefined \def \shownote      #1{#1}          \fi
\ifx \showarticletitle \undefined \def \showarticletitle #1{#1}   \fi
\ifx \showURL      \undefined \def \showURL       {\relax}        \fi
\providecommand\bibfield[2]{#2}
\providecommand\bibinfo[2]{#2}
\providecommand\natexlab[1]{#1}
\providecommand\showeprint[2][]{arXiv:#2}

\bibitem[Aiyappa et~al\mbox{.}(2023)]%
        {aiyappa2023can}
\bibfield{author}{\bibinfo{person}{Rachith Aiyappa}, \bibinfo{person}{Jisun An}, \bibinfo{person}{Haewoon Kwak}, {and} \bibinfo{person}{Yong-Yeol Ahn}.} \bibinfo{year}{2023}\natexlab{}.
\newblock \showarticletitle{Can we trust the evaluation on ChatGPT?}
\newblock \bibinfo{journal}{\emph{arXiv preprint arXiv:2303.12767}} (\bibinfo{year}{2023}).
\newblock


\bibitem[Akkaya et~al\mbox{.}(2010)]%
        {akkaya2010amazon}
\bibfield{author}{\bibinfo{person}{Cem Akkaya}, \bibinfo{person}{Alexander Conrad}, \bibinfo{person}{Janyce Wiebe}, {and} \bibinfo{person}{Rada Mihalcea}.} \bibinfo{year}{2010}\natexlab{}.
\newblock \showarticletitle{Amazon mechanical turk for subjectivity word sense disambiguation}. In \bibinfo{booktitle}{\emph{Proceedings of the NAACL HLT 2010 workshop on creating speech and language data with Amazon’s Mechanical Turk}}. \bibinfo{pages}{195--203}.
\newblock


\bibitem[Alharbi et~al\mbox{.}(2021)]%
        {alharbi2021social}
\bibfield{author}{\bibinfo{person}{Ahmed Alharbi}, \bibinfo{person}{Hai Dong}, \bibinfo{person}{Xun Yi}, \bibinfo{person}{Zahir Tari}, {and} \bibinfo{person}{Ibrahim Khalil}.} \bibinfo{year}{2021}\natexlab{}.
\newblock \showarticletitle{Social media identity deception detection: a survey}.
\newblock \bibinfo{journal}{\emph{ACM computing surveys (CSUR)}} \bibinfo{volume}{54}, \bibinfo{number}{3} (\bibinfo{year}{2021}), \bibinfo{pages}{1--35}.
\newblock


\bibitem[Ayd{\i}n and Karaarslan(2022)]%
        {aydin2022openai}
\bibfield{author}{\bibinfo{person}{{\"O}mer Ayd{\i}n} {and} \bibinfo{person}{Enis Karaarslan}.} \bibinfo{year}{2022}\natexlab{}.
\newblock \showarticletitle{OpenAI ChatGPT generated literature review: Digital twin in healthcare}.
\newblock \bibinfo{journal}{\emph{Available at SSRN 4308687}} (\bibinfo{year}{2022}).
\newblock


\bibitem[Bang et~al\mbox{.}(2023)]%
        {bang2023multitask}
\bibfield{author}{\bibinfo{person}{Yejin Bang}, \bibinfo{person}{Samuel Cahyawijaya}, \bibinfo{person}{Nayeon Lee}, \bibinfo{person}{Wenliang Dai}, \bibinfo{person}{Dan Su}, \bibinfo{person}{Bryan Wilie}, \bibinfo{person}{Holy Lovenia}, \bibinfo{person}{Ziwei Ji}, \bibinfo{person}{Tiezheng Yu}, \bibinfo{person}{Willy Chung}, {et~al\mbox{.}}} \bibinfo{year}{2023}\natexlab{}.
\newblock \showarticletitle{A multitask, multilingual, multimodal evaluation of chatgpt on reasoning, hallucination, and interactivity}.
\newblock \bibinfo{journal}{\emph{arXiv preprint arXiv:2302.04023}} (\bibinfo{year}{2023}).
\newblock


\bibitem[Barak(2005)]%
        {barak2005sexual}
\bibfield{author}{\bibinfo{person}{Azy Barak}.} \bibinfo{year}{2005}\natexlab{}.
\newblock \showarticletitle{Sexual harassment on the Internet}.
\newblock \bibinfo{journal}{\emph{Social science computer review}} \bibinfo{volume}{23}, \bibinfo{number}{1} (\bibinfo{year}{2005}), \bibinfo{pages}{77--92}.
\newblock


\bibitem[Brown et~al\mbox{.}(2020)]%
        {brown2020language}
\bibfield{author}{\bibinfo{person}{Tom Brown}, \bibinfo{person}{Benjamin Mann}, \bibinfo{person}{Nick Ryder}, \bibinfo{person}{Melanie Subbiah}, \bibinfo{person}{Jared~D Kaplan}, \bibinfo{person}{Prafulla Dhariwal}, \bibinfo{person}{Arvind Neelakantan}, \bibinfo{person}{Pranav Shyam}, \bibinfo{person}{Girish Sastry}, \bibinfo{person}{Amanda Askell}, {et~al\mbox{.}}} \bibinfo{year}{2020}\natexlab{}.
\newblock \showarticletitle{Language models are few-shot learners}.
\newblock \bibinfo{journal}{\emph{Advances in neural information processing systems}}  \bibinfo{volume}{33} (\bibinfo{year}{2020}), \bibinfo{pages}{1877--1901}.
\newblock


\bibitem[Chakraborty et~al\mbox{.}(2016)]%
        {chakraborty2016stop}
\bibfield{author}{\bibinfo{person}{Abhijnan Chakraborty}, \bibinfo{person}{Bhargavi Paranjape}, \bibinfo{person}{Sourya Kakarla}, {and} \bibinfo{person}{Niloy Ganguly}.} \bibinfo{year}{2016}\natexlab{}.
\newblock \showarticletitle{Stop clickbait: Detecting and preventing clickbaits in online news media}. In \bibinfo{booktitle}{\emph{2016 IEEE/ACM international conference on advances in social networks analysis and mining (ASONAM)}}. IEEE, \bibinfo{pages}{9--16}.
\newblock


\bibitem[D{\'\i}az et~al\mbox{.}(2022)]%
        {diaz2022crowdworksheets}
\bibfield{author}{\bibinfo{person}{Mark D{\'\i}az}, \bibinfo{person}{Ian Kivlichan}, \bibinfo{person}{Rachel Rosen}, \bibinfo{person}{Dylan Baker}, \bibinfo{person}{Razvan Amironesei}, \bibinfo{person}{Vinodkumar Prabhakaran}, {and} \bibinfo{person}{Emily Denton}.} \bibinfo{year}{2022}\natexlab{}.
\newblock \showarticletitle{Crowdworksheets: Accounting for individual and collective identities underlying crowdsourced dataset annotation}. In \bibinfo{booktitle}{\emph{Proceedings of the 2022 ACM Conference on Fairness, Accountability, and Transparency}}. \bibinfo{pages}{2342--2351}.
\newblock


\bibitem[ElSherief et~al\mbox{.}(2021)]%
        {elsherief2021latent}
\bibfield{author}{\bibinfo{person}{Mai ElSherief}, \bibinfo{person}{Caleb Ziems}, \bibinfo{person}{David Muchlinski}, \bibinfo{person}{Vaishnavi Anupindi}, \bibinfo{person}{Jordyn Seybolt}, \bibinfo{person}{Munmun De~Choudhury}, {and} \bibinfo{person}{Diyi Yang}.} \bibinfo{year}{2021}\natexlab{}.
\newblock \showarticletitle{Latent hatred: A benchmark for understanding implicit hate speech}.
\newblock \bibinfo{journal}{\emph{arXiv preprint arXiv:2109.05322}} (\bibinfo{year}{2021}).
\newblock


\bibitem[Fagni et~al\mbox{.}(2021)]%
        {fagni2021tweepfake}
\bibfield{author}{\bibinfo{person}{Tiziano Fagni}, \bibinfo{person}{Fabrizio Falchi}, \bibinfo{person}{Margherita Gambini}, \bibinfo{person}{Antonio Martella}, {and} \bibinfo{person}{Maurizio Tesconi}.} \bibinfo{year}{2021}\natexlab{}.
\newblock \showarticletitle{TweepFake: About detecting deepfake tweets}.
\newblock \bibinfo{journal}{\emph{Plos one}} \bibinfo{volume}{16}, \bibinfo{number}{5} (\bibinfo{year}{2021}), \bibinfo{pages}{e0251415}.
\newblock


\bibitem[Gilardi et~al\mbox{.}(2023)]%
        {gilardi2023chatgpt}
\bibfield{author}{\bibinfo{person}{Fabrizio Gilardi}, \bibinfo{person}{Meysam Alizadeh}, {and} \bibinfo{person}{Ma{\"e}l Kubli}.} \bibinfo{year}{2023}\natexlab{}.
\newblock \showarticletitle{Chatgpt outperforms crowd-workers for text-annotation tasks}.
\newblock \bibinfo{journal}{\emph{arXiv preprint arXiv:2303.15056}} (\bibinfo{year}{2023}).
\newblock


\bibitem[Glandt et~al\mbox{.}(2021)]%
        {glandt2021stance}
\bibfield{author}{\bibinfo{person}{Kyle Glandt}, \bibinfo{person}{Sarthak Khanal}, \bibinfo{person}{Yingjie Li}, \bibinfo{person}{Doina Caragea}, {and} \bibinfo{person}{Cornelia Caragea}.} \bibinfo{year}{2021}\natexlab{}.
\newblock \showarticletitle{Stance detection in COVID-19 tweets}. In \bibinfo{booktitle}{\emph{Proceedings of the 59th Annual Meeting of the Association for Computational Linguistics and the 11th International Joint Conference on Natural Language Processing (Long Papers)}}, Vol.~\bibinfo{volume}{1}.
\newblock


\bibitem[Guo et~al\mbox{.}(2023)]%
        {guo2023close}
\bibfield{author}{\bibinfo{person}{Biyang Guo}, \bibinfo{person}{Xin Zhang}, \bibinfo{person}{Ziyuan Wang}, \bibinfo{person}{Minqi Jiang}, \bibinfo{person}{Jinran Nie}, \bibinfo{person}{Yuxuan Ding}, \bibinfo{person}{Jianwei Yue}, {and} \bibinfo{person}{Yupeng Wu}.} \bibinfo{year}{2023}\natexlab{}.
\newblock \showarticletitle{How Close is ChatGPT to Human Experts? Comparison Corpus, Evaluation, and Detection}.
\newblock \bibinfo{journal}{\emph{arXiv preprint arXiv:2301.07597}} (\bibinfo{year}{2023}).
\newblock


\bibitem[Haq et~al\mbox{.}(2022a)]%
        {haq2022its}
\bibfield{author}{\bibinfo{person}{Ehsan-Ul Haq}, \bibinfo{person}{Yang~K. Lu}, {and} \bibinfo{person}{Pan Hui}.} \bibinfo{year}{2022}\natexlab{a}.
\newblock \showarticletitle{It's All Relative! A Method to Counter Human Bias in Crowdsourced Stance Detection of News Articles}.
\newblock  \bibinfo{volume}{6}, \bibinfo{number}{CSCW2}, Article \bibinfo{articleno}{523} (\bibinfo{date}{nov} \bibinfo{year}{2022}), \bibinfo{numpages}{25}~pages.
\newblock
\urldef\tempurl%
\url{https://doi.org/10.1145/3555636}
\showDOI{\tempurl}


\bibitem[Haq et~al\mbox{.}(2022b)]%
        {haq2022twitter}
\bibfield{author}{\bibinfo{person}{Ehsan-Ul Haq}, \bibinfo{person}{Gareth Tyson}, \bibinfo{person}{Lik-Hang Lee}, \bibinfo{person}{Tristan Braud}, {and} \bibinfo{person}{Pan Hui}.} \bibinfo{year}{2022}\natexlab{b}.
\newblock \showarticletitle{Twitter dataset for 2022 russo-ukrainian crisis}.
\newblock \bibinfo{journal}{\emph{arXiv preprint arXiv:2203.02955}} (\bibinfo{year}{2022}).
\newblock


\bibitem[He et~al\mbox{.}(2021)]%
        {he2021racism}
\bibfield{author}{\bibinfo{person}{Bing He}, \bibinfo{person}{Caleb Ziems}, \bibinfo{person}{Sandeep Soni}, \bibinfo{person}{Naren Ramakrishnan}, \bibinfo{person}{Diyi Yang}, {and} \bibinfo{person}{Srijan Kumar}.} \bibinfo{year}{2021}\natexlab{}.
\newblock \showarticletitle{Racism is a virus: Anti-Asian hate and counterspeech in social media during the COVID-19 crisis}. In \bibinfo{booktitle}{\emph{Proceedings of the 2021 IEEE/ACM International Conference on Advances in Social Networks Analysis and Mining}}. \bibinfo{pages}{90--94}.
\newblock


\bibitem[Hoes et~al\mbox{.}(2023)]%
        {hoes2023using}
\bibfield{author}{\bibinfo{person}{Emma Hoes}, \bibinfo{person}{Sacha Altay}, {and} \bibinfo{person}{Juan Bermeo}.} \bibinfo{year}{2023}\natexlab{}.
\newblock \showarticletitle{Using ChatGPT to Fight Misinformation: ChatGPT Nails 72\% of 12,000 Verified Claims}.
\newblock  (\bibinfo{year}{2023}).
\newblock


\bibitem[Huang et~al\mbox{.}(2023)]%
        {huang2023chatgpt}
\bibfield{author}{\bibinfo{person}{Fan Huang}, \bibinfo{person}{Haewoon Kwak}, {and} \bibinfo{person}{Jisun An}.} \bibinfo{year}{2023}\natexlab{}.
\newblock \showarticletitle{Is ChatGPT better than Human Annotators? Potential and Limitations of ChatGPT in Explaining Implicit Hate Speech}.
\newblock \bibinfo{journal}{\emph{arXiv preprint arXiv:2302.07736}} (\bibinfo{year}{2023}).
\newblock


\bibitem[Kennedy et~al\mbox{.}(2020)]%
        {kennedy2020constructing}
\bibfield{author}{\bibinfo{person}{Chris~J Kennedy}, \bibinfo{person}{Geoff Bacon}, \bibinfo{person}{Alexander Sahn}, {and} \bibinfo{person}{Claudia von Vacano}.} \bibinfo{year}{2020}\natexlab{}.
\newblock \showarticletitle{Constructing interval variables via faceted Rasch measurement and multitask deep learning: a hate speech application}.
\newblock \bibinfo{journal}{\emph{arXiv preprint arXiv:2009.10277}} (\bibinfo{year}{2020}).
\newblock


\bibitem[Kuzman et~al\mbox{.}(2023)]%
        {kuzman2023chatgpt}
\bibfield{author}{\bibinfo{person}{Taja Kuzman}, \bibinfo{person}{Igor Mozetic}, {and} \bibinfo{person}{Nikola Ljube{\v{s}}ic}.} \bibinfo{year}{2023}\natexlab{}.
\newblock \showarticletitle{ChatGPT: Beginning of an End of Manual Linguistic Data Annotation? Use Case of Automatic Genre Identification}.
\newblock \bibinfo{journal}{\emph{arXiv e-prints}} (\bibinfo{year}{2023}), \bibinfo{pages}{arXiv--2303}.
\newblock


\bibitem[Lee et~al\mbox{.}(2011)]%
        {lee2011twitter}
\bibfield{author}{\bibinfo{person}{Kathy Lee}, \bibinfo{person}{Diana Palsetia}, \bibinfo{person}{Ramanathan Narayanan}, \bibinfo{person}{Md~Mostofa~Ali Patwary}, \bibinfo{person}{Ankit Agrawal}, {and} \bibinfo{person}{Alok Choudhary}.} \bibinfo{year}{2011}\natexlab{}.
\newblock \showarticletitle{Twitter trending topic classification}. In \bibinfo{booktitle}{\emph{2011 IEEE 11th international conference on data mining workshops}}. IEEE, \bibinfo{pages}{251--258}.
\newblock


\bibitem[Li et~al\mbox{.}(2021)]%
        {li2021p}
\bibfield{author}{\bibinfo{person}{Yingjie Li}, \bibinfo{person}{Tiberiu Sosea}, \bibinfo{person}{Aditya Sawant}, \bibinfo{person}{Ajith~Jayaraman Nair}, \bibinfo{person}{Diana Inkpen}, {and} \bibinfo{person}{Cornelia Caragea}.} \bibinfo{year}{2021}\natexlab{}.
\newblock \showarticletitle{P-stance: A large dataset for stance detection in political domain}. In \bibinfo{booktitle}{\emph{Findings of the Association for Computational Linguistics: ACL-IJCNLP 2021}}. \bibinfo{pages}{2355--2365}.
\newblock


\bibitem[Mathew et~al\mbox{.}(2019)]%
        {mathew2019thou}
\bibfield{author}{\bibinfo{person}{Binny Mathew}, \bibinfo{person}{Punyajoy Saha}, \bibinfo{person}{Hardik Tharad}, \bibinfo{person}{Subham Rajgaria}, \bibinfo{person}{Prajwal Singhania}, \bibinfo{person}{Suman~Kalyan Maity}, \bibinfo{person}{Pawan Goyal}, {and} \bibinfo{person}{Animesh Mukherjee}.} \bibinfo{year}{2019}\natexlab{}.
\newblock \showarticletitle{Thou shalt not hate: Countering online hate speech}. In \bibinfo{booktitle}{\emph{Proceedings of the international AAAI conference on web and social media}}, Vol.~\bibinfo{volume}{13}. \bibinfo{pages}{369--380}.
\newblock


\bibitem[Mohammad et~al\mbox{.}(2016)]%
        {mohammad2016semeval}
\bibfield{author}{\bibinfo{person}{Saif Mohammad}, \bibinfo{person}{Svetlana Kiritchenko}, \bibinfo{person}{Parinaz Sobhani}, \bibinfo{person}{Xiaodan Zhu}, {and} \bibinfo{person}{Colin Cherry}.} \bibinfo{year}{2016}\natexlab{}.
\newblock \showarticletitle{Semeval-2016 task 6: Detecting stance in tweets}. In \bibinfo{booktitle}{\emph{Proceedings of the 10th international workshop on semantic evaluation (SemEval-2016)}}. \bibinfo{pages}{31--41}.
\newblock


\bibitem[Nadeau and Sekine(2007)]%
        {nadeau2007survey}
\bibfield{author}{\bibinfo{person}{David Nadeau} {and} \bibinfo{person}{Satoshi Sekine}.} \bibinfo{year}{2007}\natexlab{}.
\newblock \showarticletitle{A survey of named entity recognition and classification}.
\newblock \bibinfo{journal}{\emph{Lingvisticae Investigationes}} \bibinfo{volume}{30}, \bibinfo{number}{1} (\bibinfo{year}{2007}), \bibinfo{pages}{3--26}.
\newblock


\bibitem[Neelakantan et~al\mbox{.}(2022)]%
        {neelakantan2022text}
\bibfield{author}{\bibinfo{person}{Arvind Neelakantan}, \bibinfo{person}{Tao Xu}, \bibinfo{person}{Raul Puri}, \bibinfo{person}{Alec Radford}, \bibinfo{person}{Jesse~Michael Han}, \bibinfo{person}{Jerry Tworek}, \bibinfo{person}{Qiming Yuan}, \bibinfo{person}{Nikolas Tezak}, \bibinfo{person}{Jong~Wook Kim}, \bibinfo{person}{Chris Hallacy}, {et~al\mbox{.}}} \bibinfo{year}{2022}\natexlab{}.
\newblock \showarticletitle{Text and code embeddings by contrastive pre-training}.
\newblock \bibinfo{journal}{\emph{arXiv preprint arXiv:2201.10005}} (\bibinfo{year}{2022}).
\newblock


\bibitem[Patwa et~al\mbox{.}(2020)]%
        {patwa2020fighting}
\bibfield{author}{\bibinfo{person}{Parth Patwa}, \bibinfo{person}{Shivam Sharma}, \bibinfo{person}{Srinivas PYKL}, \bibinfo{person}{Vineeth Guptha}, \bibinfo{person}{Gitanjali Kumari}, \bibinfo{person}{Md~Shad Akhtar}, \bibinfo{person}{Asif Ekbal}, \bibinfo{person}{Amitava Das}, {and} \bibinfo{person}{Tanmoy Chakraborty}.} \bibinfo{year}{2020}\natexlab{}.
\newblock \bibinfo{title}{Fighting an Infodemic: COVID-19 Fake News Dataset}.
\newblock
\newblock
\showeprint[arxiv]{2011.03327}~[cs.CL]


\bibitem[Peixian et~al\mbox{.}(2023)]%
        {RU-echo}
\bibfield{author}{\bibinfo{person}{Zhang Peixian}, \bibinfo{person}{Haq Ehsan-Ul}, \bibinfo{person}{Zhu Yiming}, \bibinfo{person}{Hui Pan}, {and} \bibinfo{person}{Tyson Gareth}.} \bibinfo{year}{2023}\natexlab{}.
\newblock \showarticletitle{Echo Chambers within the Russo-Ukrainian War: The Role of Bipartisan Users.}. In \bibinfo{booktitle}{\emph{2023 IEEE/ACM International Conference on Advances in Social Networks Analysis and Mining (ASONAM)}}.
\newblock
\urldef\tempurl%
\url{https://doi.org/10.1145/3625007.3627475}
\showDOI{\tempurl}


\bibitem[Poddar et~al\mbox{.}(2022)]%
        {poddar2022winds}
\bibfield{author}{\bibinfo{person}{Soham Poddar}, \bibinfo{person}{Mainack Mondal}, \bibinfo{person}{Janardan Misra}, \bibinfo{person}{Niloy Ganguly}, {and} \bibinfo{person}{Saptarshi Ghosh}.} \bibinfo{year}{2022}\natexlab{}.
\newblock \showarticletitle{Winds of Change: Impact of COVID-19 on Vaccine-related Opinions of Twitter users}. In \bibinfo{booktitle}{\emph{Proceedings of the International AAAI Conference on Web and Social Media}}, Vol.~\bibinfo{volume}{16}. \bibinfo{pages}{782--793}.
\newblock


\bibitem[Rosenthal et~al\mbox{.}(2019)]%
        {rosenthal2019semeval}
\bibfield{author}{\bibinfo{person}{Sara Rosenthal}, \bibinfo{person}{Noura Farra}, {and} \bibinfo{person}{Preslav Nakov}.} \bibinfo{year}{2019}\natexlab{}.
\newblock \showarticletitle{SemEval-2017 task 4: Sentiment analysis in Twitter}.
\newblock \bibinfo{journal}{\emph{arXiv preprint arXiv:1912.00741}} (\bibinfo{year}{2019}).
\newblock


\bibitem[Sallam et~al\mbox{.}(2023)]%
        {sallam2023chatgpt}
\bibfield{author}{\bibinfo{person}{Malik Sallam}, \bibinfo{person}{Nesreen~A Salim}, \bibinfo{person}{B Ala’a}, \bibinfo{person}{Muna Barakat}, \bibinfo{person}{Diaa Fayyad}, \bibinfo{person}{Souheil Hallit}, \bibinfo{person}{Harapan Harapan}, \bibinfo{person}{Rabih Hallit}, \bibinfo{person}{Azmi Mahafzah}, {and} \bibinfo{person}{B Ala'a}.} \bibinfo{year}{2023}\natexlab{}.
\newblock \showarticletitle{ChatGPT Output Regarding Compulsory Vaccination and COVID-19 Vaccine Conspiracy: A Descriptive Study at the Outset of a Paradigm Shift in Online Search for Information}.
\newblock \bibinfo{journal}{\emph{Cureus}} \bibinfo{volume}{15}, \bibinfo{number}{2} (\bibinfo{year}{2023}).
\newblock


\bibitem[Schieb and Preuss(2016)]%
        {schieb2016governing}
\bibfield{author}{\bibinfo{person}{Carla Schieb} {and} \bibinfo{person}{Mike Preuss}.} \bibinfo{year}{2016}\natexlab{}.
\newblock \showarticletitle{Governing hate speech by means of counterspeech on Facebook}. In \bibinfo{booktitle}{\emph{66th ica annual conference, at fukuoka, japan}}. \bibinfo{pages}{1--23}.
\newblock


\bibitem[Sera et~al\mbox{.}(2002)]%
        {sera2002language}
\bibfield{author}{\bibinfo{person}{Maria~D Sera}, \bibinfo{person}{Chryle Elieff}, \bibinfo{person}{James Forbes}, \bibinfo{person}{Melissa~Clark Burch}, \bibinfo{person}{Wanda Rodr{\'\i}guez}, {and} \bibinfo{person}{Diane~Poulin Dubois}.} \bibinfo{year}{2002}\natexlab{}.
\newblock \showarticletitle{When language affects cognition and when it does not: An analysis of grammatical gender and classification.}
\newblock \bibinfo{journal}{\emph{Journal of Experimental Psychology: General}} \bibinfo{volume}{131}, \bibinfo{number}{3} (\bibinfo{year}{2002}), \bibinfo{pages}{377}.
\newblock


\bibitem[Shu et~al\mbox{.}(2017)]%
        {shu2017fake}
\bibfield{author}{\bibinfo{person}{Kai Shu}, \bibinfo{person}{Amy Sliva}, \bibinfo{person}{Suhang Wang}, \bibinfo{person}{Jiliang Tang}, {and} \bibinfo{person}{Huan Liu}.} \bibinfo{year}{2017}\natexlab{}.
\newblock \showarticletitle{Fake news detection on social media: A data mining perspective}.
\newblock \bibinfo{journal}{\emph{ACM SIGKDD explorations newsletter}} \bibinfo{volume}{19}, \bibinfo{number}{1} (\bibinfo{year}{2017}), \bibinfo{pages}{22--36}.
\newblock


\bibitem[Snow et~al\mbox{.}(2008)]%
        {snow2008cheap}
\bibfield{author}{\bibinfo{person}{Rion Snow}, \bibinfo{person}{Brendan O’connor}, \bibinfo{person}{Dan Jurafsky}, {and} \bibinfo{person}{Andrew~Y Ng}.} \bibinfo{year}{2008}\natexlab{}.
\newblock \showarticletitle{Cheap and fast--but is it good? evaluating non-expert annotations for natural language tasks}. In \bibinfo{booktitle}{\emph{Proceedings of the 2008 conference on empirical methods in natural language processing}}. \bibinfo{pages}{254--263}.
\newblock


\bibitem[Sobania et~al\mbox{.}(2023)]%
        {sobania2023analysis}
\bibfield{author}{\bibinfo{person}{Dominik Sobania}, \bibinfo{person}{Martin Briesch}, \bibinfo{person}{Carol Hanna}, {and} \bibinfo{person}{Justyna Petke}.} \bibinfo{year}{2023}\natexlab{}.
\newblock \showarticletitle{An analysis of the automatic bug fixing performance of chatgpt}.
\newblock \bibinfo{journal}{\emph{arXiv preprint arXiv:2301.08653}} (\bibinfo{year}{2023}).
\newblock


\bibitem[Sorokin and Forsyth(2008)]%
        {sorokin2009utility}
\bibfield{author}{\bibinfo{person}{Alexander Sorokin} {and} \bibinfo{person}{David Forsyth}.} \bibinfo{year}{2008}\natexlab{}.
\newblock \showarticletitle{Utility data annotation with Amazon Mechanical Turk}. In \bibinfo{booktitle}{\emph{2008 IEEE Computer Society Conference on Computer Vision and Pattern Recognition Workshops}}. \bibinfo{pages}{1--8}.
\newblock
\urldef\tempurl%
\url{https://doi.org/10.1109/CVPRW.2008.4562953}
\showDOI{\tempurl}


\bibitem[ul~Haq et~al\mbox{.}(2020)]%
        {ul2020enemy}
\bibfield{author}{\bibinfo{person}{Ehsan ul Haq}, \bibinfo{person}{Tristan Braud}, \bibinfo{person}{Young~D Kwon}, {and} \bibinfo{person}{Pan Hui}.} \bibinfo{year}{2020}\natexlab{}.
\newblock \showarticletitle{Enemy at the Gate: Evolution of Twitter User's Polarization During National Crisis}. In \bibinfo{booktitle}{\emph{2020 IEEE/ACM International Conference on Advances in Social Networks Analysis and Mining (ASONAM)}}. IEEE, \bibinfo{pages}{212--216}.
\newblock


\bibitem[Usama et~al\mbox{.}(2019)]%
        {usama2019unsupervised}
\bibfield{author}{\bibinfo{person}{Muhammad Usama}, \bibinfo{person}{Junaid Qadir}, \bibinfo{person}{Aunn Raza}, \bibinfo{person}{Hunain Arif}, \bibinfo{person}{Kok-Lim~Alvin Yau}, \bibinfo{person}{Yehia Elkhatib}, \bibinfo{person}{Amir Hussain}, {and} \bibinfo{person}{Ala Al-Fuqaha}.} \bibinfo{year}{2019}\natexlab{}.
\newblock \showarticletitle{Unsupervised machine learning for networking: Techniques, applications and research challenges}.
\newblock \bibinfo{journal}{\emph{IEEE access}}  \bibinfo{volume}{7} (\bibinfo{year}{2019}), \bibinfo{pages}{65579--65615}.
\newblock


\bibitem[Wang et~al\mbox{.}(2023)]%
        {wang2023chatgpt}
\bibfield{author}{\bibinfo{person}{Jiaan Wang}, \bibinfo{person}{Yunlong Liang}, \bibinfo{person}{Fandong Meng}, \bibinfo{person}{Haoxiang Shi}, \bibinfo{person}{Zhixu Li}, \bibinfo{person}{Jinan Xu}, \bibinfo{person}{Jianfeng Qu}, {and} \bibinfo{person}{Jie Zhou}.} \bibinfo{year}{2023}\natexlab{}.
\newblock \showarticletitle{Is chatgpt a good nlg evaluator? a preliminary study}.
\newblock \bibinfo{journal}{\emph{arXiv preprint arXiv:2303.04048}} (\bibinfo{year}{2023}).
\newblock


\bibitem[Wang et~al\mbox{.}(2020)]%
        {wang2020survey}
\bibfield{author}{\bibinfo{person}{Meng Wang}, \bibinfo{person}{Weijie Fu}, \bibinfo{person}{Xiangnan He}, \bibinfo{person}{Shijie Hao}, {and} \bibinfo{person}{Xindong Wu}.} \bibinfo{year}{2020}\natexlab{}.
\newblock \showarticletitle{A survey on large-scale machine learning}.
\newblock \bibinfo{journal}{\emph{IEEE Transactions on Knowledge and Data Engineering}} \bibinfo{volume}{34}, \bibinfo{number}{6} (\bibinfo{year}{2020}), \bibinfo{pages}{2574--2594}.
\newblock


\bibitem[Wei et~al\mbox{.}(2022)]%
        {wei2022chain}
\bibfield{author}{\bibinfo{person}{Jason Wei}, \bibinfo{person}{Xuezhi Wang}, \bibinfo{person}{Dale Schuurmans}, \bibinfo{person}{Maarten Bosma}, \bibinfo{person}{Fei Xia}, \bibinfo{person}{Ed Chi}, \bibinfo{person}{Quoc~V Le}, \bibinfo{person}{Denny Zhou}, {et~al\mbox{.}}} \bibinfo{year}{2022}\natexlab{}.
\newblock \showarticletitle{Chain-of-thought prompting elicits reasoning in large language models}.
\newblock \bibinfo{journal}{\emph{Advances in Neural Information Processing Systems}}  \bibinfo{volume}{35} (\bibinfo{year}{2022}), \bibinfo{pages}{24824--24837}.
\newblock


\bibitem[Zhang et~al\mbox{.}(2022)]%
        {zhang2022would}
\bibfield{author}{\bibinfo{person}{Bowen Zhang}, \bibinfo{person}{Daijun Ding}, {and} \bibinfo{person}{Liwen Jing}.} \bibinfo{year}{2022}\natexlab{}.
\newblock \showarticletitle{How would Stance Detection Techniques Evolve after the Launch of ChatGPT?}
\newblock \bibinfo{journal}{\emph{arXiv preprint arXiv:2212.14548}} (\bibinfo{year}{2022}).
\newblock


\bibitem[Zhang et~al\mbox{.}(2023)]%
        {zhang2023complete}
\bibfield{author}{\bibinfo{person}{Chaoning Zhang}, \bibinfo{person}{Chenshuang Zhang}, \bibinfo{person}{Sheng Zheng}, \bibinfo{person}{Yu Qiao}, \bibinfo{person}{Chenghao Li}, \bibinfo{person}{Mengchun Zhang}, \bibinfo{person}{Sumit~Kumar Dam}, \bibinfo{person}{Chu~Myaet Thwal}, \bibinfo{person}{Ye~Lin Tun}, \bibinfo{person}{Le~Luang Huy}, {et~al\mbox{.}}} \bibinfo{year}{2023}\natexlab{}.
\newblock \showarticletitle{A Complete Survey on Generative AI (AIGC): Is ChatGPT from GPT-4 to GPT-5 All You Need?}
\newblock \bibinfo{journal}{\emph{arXiv preprint arXiv:2303.11717}} (\bibinfo{year}{2023}).
\newblock


\bibitem[Zhou and Zafarani(2020)]%
        {zhou2020survey}
\bibfield{author}{\bibinfo{person}{Xinyi Zhou} {and} \bibinfo{person}{Reza Zafarani}.} \bibinfo{year}{2020}\natexlab{}.
\newblock \showarticletitle{A survey of fake news: Fundamental theories, detection methods, and opportunities}.
\newblock \bibinfo{journal}{\emph{ACM Computing Surveys (CSUR)}} \bibinfo{volume}{53}, \bibinfo{number}{5} (\bibinfo{year}{2020}), \bibinfo{pages}{1--40}.
\newblock


\bibitem[Zhu et~al\mbox{.}(2022)]%
        {zhu2022reddit}
\bibfield{author}{\bibinfo{person}{Yiming Zhu}, \bibinfo{person}{Ehsan-ul Haq}, \bibinfo{person}{Lik-Hang Lee}, \bibinfo{person}{Gareth Tyson}, {and} \bibinfo{person}{Pan Hui}.} \bibinfo{year}{2022}\natexlab{}.
\newblock \showarticletitle{A Reddit Dataset for the Russo-Ukrainian Conflict in 2022}.
\newblock \bibinfo{journal}{\emph{arXiv preprint arXiv:2206.05107}} (\bibinfo{year}{2022}).
\newblock


\end{thebibliography}










\end{document}